# A FRAMEWORK FOR INTELLIGENT MEDICAL DIAGNOSIS USING ROUGH SET WITH FORMAL CONCEPT ANALYSIS

B. K. Tripathy[1], D. P. Acharjya[1] and V. Cynthya[1]

[1]School of Computing Science and Engineering, VIT University, Vellore, India
tripathybk@rediffmail.com, dpacharjya@gmail.com,
cynthya.isaacs1987@gmail.com

## ABSTRACT

*Medical diagnosis process vary in the degree to which they attempt to deal with different complicating aspects of diagnosis such as relative importance of symptoms, varied symptom pattern and the relation between diseases them selves. Based on decision theory, in the past many mathematical models such as crisp set, probability distribution, fuzzy set, intuitionistic fuzzy set were developed to deal with complicating aspects of diagnosis. But, many such models are failed to include important aspects of the expert decisions. Therefore, an effort has been made to process inconsistencies in data being considered by Pawlak with the introduction of rough set theory. Though rough set has major advantages over the other methods, but it generates too many rules that create many difficulties while taking decisions. Therefore, it is essential to minimize the decision rules. In this paper, we use two processes such as pre process and post process to mine suitable rules and to explore the relationship among the attributes. In pre process we use rough set theory to mine suitable rules, whereas in post process we use formal concept analysis from these suitable rules to explore better knowledge and most important factors affecting the decision making.*

## 1. INTRODUCTION

Now-a-days Internet is the best example for distributed computing which involves dispersion of data geographically. Therefore, it is challenging for human being to retrieve information from the huge amount of data available geographically at different health research centres. Hence, it is very difficult to extract expert knowledge from the universe of medical dataset. The problem of imperfect knowledge has been tackled for a long time by philosophers, logicians, and mathematicians. Recently it brings attention for computer scientists, particularly in the area of knowledge mining and artificial intelligence. There are many approaches to the problem of how to understand and manipulate imperfect knowledge. The fundamental one is the crisp set. However, it has been extended in many directions as far as modelling of real life situations are concerned. The earliest and most successful one is being the notion of fuzzy sets by L. A. Zadeh [1] that captures impreciseness in information. On the other hand rough sets of Z. Pawlak [2] is another attempt that capture indeiscernibility among objects to model imperfect knowledge [3, 4, 5]. There were many other advanced methods such as rough set with similarity, fuzzy rough set, rough set on fuzzy approximation spaces, rough set intuitionistic fuzzy approximation spaces, dynamic rough set, covering based rough set were discussed by different authors to extract knowledge from the huge amount of data [6, 7, 8, 9, 10, 11]. Universe can be considered as a large collection of objects. Each object is associated with some information with it. In order to find knowledge about the universe we need to extract some information about these objects. We need sufficient amount of information to uniquely identify the objects which is not possible in case of all objects. Therefore, we require classification of these objects into similarity classes to characterize these objects in order to extract knowledge about the universe. Rough set is an approach to extract knowledge and association between data and values of data in recent years.

However, it generates too many rules that create many difficulties in taking decision for human being. Hence it is challenging for human being to extract expert knowledge. However, many researchers has analysed medical data by using data mining, fuzzy sets, and formal concept analysis for finding decision rules, and redundancies[12, 13].

In this paper, we use two processes such as pre process and post process to mine suitable rules and to explore the relationship among the attributes. In pre process we use rough set to mine suitable rules, whereas in post process we use formal concept analysis to explore better knowledge and most important characteristics affecting the decision making. The remainder of the paper is organized as follows: Section 2 presents the basics of rough set theory. Section 3 provides the basic idea of formal concept analysis. The proposed intelligent mining model is given in Sections 4. In Section 5, an empirical study on heart disease is presented. This further followed by a conclusion in Section 6.

## 2. FOUNDATIONS OF ROUGH SET THEORY

Convergence of information and communication technologies brought a radical change in the way medical diagnosis is carried out. It is well established fact that right decision at right time provides an advantage in medical diagnosis. But, the real challenge lies in converting voluminous data into knowledge and to use this knowledge to make proper medical diagnosis. Though present technologies help in creating large databases, but most of the information may not be relevant. Therefore, attribute reduction becomes an important aspect for handling such voluminous database by eliminating superfluous data. Rough set theory developed by Z. Pawlak [14] used to process uncertain and incomplete information is a tool to the above mentioned problem. However, it has many applications in all the fields of science and engineering. One of its strength is the attribute dependencies, their significance among inconsistent data. At the same time, it does not need any preliminary or additional information about the data. Therefore, it classifies imprecise, uncertain or incomplete information expressed in terms of data.

### 2.1. Rough Sets

In this section we recall the basic definitions of basic rough set theory developed by Z. Pawlak [14]. Let *U* be a finite nonempty set called the universe. Suppose $R \subseteq U \times U$ is an equivalence relation on *U*. The equivalence relation *R* partitions the set *U* into disjoint subsets. Elements of same equivalence class are said to be indistinguishable. Equivalence classes induced by *R* are called elementary concepts. Every union of elementary concepts is called a definable set. The empty set is considered to be a definable set, thus all the definable sets form a Boolean algebra and (*U*, *R*) is called an approximation space. Given a target set *X*, we can characterize *X* by a pair of lower and upper approximations. We associate two subsets $\underline{R}X$ and $\overline{R}X$ called the *R* – lower and *R*–upper approximations of *X* respectively and are given by

$$\underline{R}X = \bigcup \{Y \in U / R : Y \subseteq X\} \qquad (1)$$

and $$\overline{R}X = \bigcup \{Y \in U / R : Y \cap X \neq \phi\} \qquad (2)$$

The *R*–boundary of *X*, $BN_R(X)$ is given by $BN_R(X) = \overline{R}X - \underline{R}X$. We say *X* is rough with respect to *R* if and only if $\overline{R}X \neq \underline{R}X$, equivalently $BN_R(\mathrm{X}) \neq \phi$. *X* is said to be *R* – definable if and only if $\overline{R}X = \underline{R}X$ or $BN_R(X) = \phi$. So, a set is rough with respect to *R* if and only if it is not *R* – definable.

### 2.2. Information System

As mentioned in the previous section rough set philosophy is based on the assumption that, in addition to crisp set theory, we have some additional information about the elements of a universe of discourse. Elements that exhibit the same information are indiscernible and form the basic building blocks that can be considered as elementary concept of knowledge about the universe.

An information system is a table that provides a convenient way to describe a finite set of objects called the universe by a finite set of attributes thereby representing all available information and knowledge. The attribute sets along with the objects in an information system consists of the set of condition attributes and decision attributes. Therefore it is also named as decision table [15]. Let us denote the information system as $I = (U, A, V, f)$, where $U$ is a finite non-empty set of objects called the universe and $A$ is a non-empty finite set of attributes. For every $a \in A$, $V_a$ is the set of values that attribute $a$ may take. Also $V = \bigcup_{a \in A} V_a$. In addition, for every $a \in A$, $f_a : U \to V_a$ is the information function. Also $f : U \times A \to V$ [16].

Consider a medical information system Table 1 of ten patients $p_i$; $i = 1, 2, \cdots, 10$ as the set of objects of the universe with a set of attributes cough, vomiting, cold severing, nasal bleeding, temperature, delirium and fever. The attribute fever is considered as the decision attribute. Let us assume that $a_1, a_2, a_3, a_4, a_5, a_6, d$ as cough, vomiting, cold severing, nasal bleeding, temperature, delirium and fever respectively. In particular, patient ($p_1$) is characterized in the table by the attribute value set (cough, always), (vomiting, seldom), (cold severing, seldom), (nasal bleeding, never), (temperature, high), (delirium, never), and (fever, yes) which form the information about the patient.

Table 1. Classical view of the medical information system

| Patient | Cough | Vomiting | Cold Severing | Nasal bleeding | Temperature | Delirium | Fever |
|---|---|---|---|---|---|---|---|
| $p_1$ | Always | Seldom | Seldom | Never | High | Never | Yes |
| $p_2$ | Seldom | Never | Always | Never | Normal | Never | No |
| $p_3$ | Never | Always | Seldom | Never | Very high | Never | Yes |
| $p_4$ | Always | Seldom | Seldom | Always | Normal | Never | No |
| $p_5$ | Never | Always | Seldom | Always | High | Never | No |
| $p_6$ | Never | Seldom | Always | Always | Normal | Never | Yes |
| $p_7$ | Always | Always | Seldom | Seldom | Normal | Never | No |
| $p_8$ | Seldom | Never | Always | Seldom | Very high | Never | Yes |
| $p_9$ | Always | Always | Never | Never | High | Never | Yes |
| $p_{10}$ | Seldom | Seldom | Seldom | Seldom | High | Never | Yes |

To make our analysis simpler we have assigned values 1, 2, 3, 4, 5, 6 to the attribute values always, seldom, never, normal, high, and very high respectively. Also, we have assigned 1, and 2 for decision values yes, no respectively. However, these values are optional and do not affect the analysis. The reduced medical information system is presented in Table 2.

Table 2. Reduced medical information system

| Patient | $a_1$ | $a_2$ | $a_3$ | $a_4$ | $a_5$ | $a_6$ | $d$ |
|---|---|---|---|---|---|---|---|
| $p_1$ | 1 | 2 | 2 | 3 | 5 | 3 | 1 |
| $p_2$ | 2 | 3 | 1 | 3 | 4 | 3 | 2 |
| $p_3$ | 3 | 1 | 2 | 3 | 6 | 3 | 1 |
| $p_4$ | 1 | 2 | 2 | 1 | 4 | 3 | 2 |
| $p_5$ | 3 | 1 | 2 | 1 | 5 | 3 | 2 |
| $p_6$ | 3 | 2 | 1 | 1 | 4 | 3 | 1 |
| $p_7$ | 1 | 1 | 2 | 2 | 4 | 3 | 2 |
| $p_8$ | 2 | 3 | 1 | 2 | 6 | 3 | 1 |
| $p_9$ | 1 | 1 | 3 | 3 | 5 | 3 | 1 |
| $p_{10}$ | 2 | 2 | 2 | 2 | 5 | 3 | 1 |

### 2.3. Indiscernibility Relation

Universe can be considered as a large collection of objects. Each object is associated with some information (data, knowledge) within it. In order to find knowledge about the universe we need to process these attribute values. Therefore, we require sufficient amount of information to uniquely identify, classify these objects into similar classes and to extract knowledge about the universe. The classification of the objects of the universe is done based on indiscernibility relation among these objects. It indicates that objects of a class cannot discern from one another based on available set of attributes of the objects [14, 17]. The indiscernibility relation generated in this way is the mathematical basis of rough set theory. Any set of all indiscernible objects is called an elementary concept, and form a basic granule (atom) of knowledge about the universe. Any union of the elementary sets is referred to be either crisp (precise) set or rough (imprecise) set. Let $P \subseteq A$ and $x_i, x_j \in U$. Then we say $x_i$ and $x_j$ are indiscernible by the set of attributes $P$ in $A$ if and only if the following (3) holds.

$$f(x_i, a) = f(x_j, a), \quad \forall a \in P \tag{3}$$

For example, given the attributes cough, vomiting, cold severing, and delirium $\{p_1, p_4\}$ are indiscernible. Similarly, the other indiscernible classes obtained are $\{p_2, p_8\}, \{p_3, p_5\}, \{p_6\}, \{p_7\}, \{p_9\}$, and $\{p_{10}\}$. In general each patient $p_i, i = 1, 2, \cdots, 10$ is compared with each other cell wise to find the indiscernibility in the attribute value. From the data set Table-1, on considering the attributes A = {cough, vomiting, cold severing, delirium}, we get the family of equivalence classes of A, i.e., the partition determined by set of attributes A, denoted by $U/\mathrm{A}$ or $I(\mathrm{A})$. Therefore,

$$U/\mathrm{A} = \{\{p_1, p_4\}, \{p_2, p_8\}, \{p_3, p_5\}, \{p_6\}, \{p_7\}, \{p_9\}, \{p_{10}\}\}$$

On considering the target set $X = \{p_1, p_6, p_9, p_{10}\}$; patients $p_1$ and $p_4$ are the boundary-line objects, where

$$\underline{\mathrm{A}}X = \bigcup\{Y \in U/\mathrm{A} : Y \subseteq X\} = \{p_6, p_9, p_{10}\} \quad \text{and}$$

$$\overline{\mathrm{A}}X = \bigcup\{Y \in U/\mathrm{A} : Y \cap X \neq \phi\} = \{p_1, p_4, p_6, p_9, p_{10}\}$$

Furthermore, considering the attributes $P \subseteq \mathrm{A}$, we can associate an index (*i.e.* $\alpha_{\mathrm{A}}(X)$) called the accuracy of approximation for any set $X \subseteq U$ as follows:

$$\alpha_{\mathrm{A}}(X) = \frac{\text{cardinality of } \underline{\mathrm{A}}\mathrm{X}}{\text{cardinality of } \overline{\mathrm{A}}X} = \frac{|\underline{\mathrm{A}}\mathrm{X}|}{|\overline{\mathrm{A}}X|}$$

For example, $\alpha_{\mathrm{A}}(X) = \frac{3}{5}$, where $X = \{p_1, p_6, p_9, p_{10}\}$; and A={cough, vomiting, cold severing, delirium}. From the information system given in Table 1, it is clear that patient $p_1$ has fever, whereas patient $p_4$ do not and they are indiscernible with respect to the attributes cough, vomiting, cold severing, and delirium. Hence, fever can not be characterized in terms of symptoms and symptoms-value pair (cough, always), (vomiting, seldom), (cold severing, seldom), and (delirium, never). Therefore, patients $p_1$ and $p_4$ are the boundary line objects and can not be classified in view of the available knowledge. On the other hand the patients $p_6$, $p_9$, and $p_{10}$ display symptoms with certainty as having fever. Therefore, the lower approximation of $X$ is $\{p_6, p_9, p_{10}\}$, whereas the upper approximation is $\{p_1, p_4, p_6, p_9, p_{10}\}$.

### 2.4. Reduct and Rule Discovery

One of the important aspects of the rough set theory is the attribute reduction and core. In an information system, some attributes may be redundant and useless. If those redundant and useless attributes are removed without affecting the classification power of attributes, we can call them the superfluous attributes [18]. The core concept is commonly used in all reducts [19].

If the set of attributes is dependent, using the dependency properties of attributes, we are interested in finding all possible minimal subsets of attributes which have the same number of elementary sets without loss of the classification power of the reduced information system [20]. In order to express the above notions more clearly we need some auxiliary notations. Let $P \subseteq A$ and $a \in P$. We say that the attribute $a$ is dispensable in $P$ if the following condition (4) holds; otherwise $a$ is indispensable in $P$.

$$I(P) = I(P - \{a\}) \tag{4}$$

Set $P$ is independent if all its attributes are indispensable. Reduct ($P'$) of $P$ is a subset of attributes $P$ such that the equivalence class induced by the reduced attribute set $P'$ are the same as the equivalence class structure induced by the attribute set $P$. *i.e.*, $I(P) = I(P')$. The core of the attribute set $P$ is the set of all indispensable attributes of $P$. The important property connecting the notion of core and reducts is defined in (5), where Red($P$) is the set of all reducts of $P$.

$$Core(P) = \bigcap \text{Red}(P) \tag{5}$$

An information system is defined as $I = (U, A, V, f)$, where $U$ is a non empty finite set of objects, $A = C \cup D$ is a non empty finite set of attributes, $C$ denotes the set of condition attributes and $D$ denotes the set of decision attributes, $C \cap D = \phi$. According to Pawlak [21], a decision rule $S$ in information system is expressed as $\varphi \to \psi$, where $\varphi$ and $\psi$ are conditions and decisions of the decision rule $S$ respectively. There are three measurements for decision rules. The first one is the accuracy of the decision rule. The second one is the support of a rule whereas the third measurement is the strength of the decision rule. The support of a rule is defined as $\text{Supp}_S(\varphi, \psi) = \text{Card}(\|\varphi \wedge \psi\|_S)$, whereas the strength of the decision rule $\varphi \to \psi$ is defined as:

$$\sigma(\varphi, \psi) = \frac{\text{Supp}_S(\varphi, \psi)}{\text{Card}(\|U\|_\psi)} \tag{6}$$

It implies that stronger rules cover more number of supporting objects and its strength can be calculated by using relation (6). The relative decision rules for Table 2 are given in Table 3, where we use $R_{i,d}$ for $i^{th}$ rule in the decision class $d$.

Table 3. Relative decision rules

| Rule Number | Notation | Decision class | Meaning | Support | Strength rate (%) | Accuracy rate (%) |
|---|---|---|---|---|---|---|
| 1 | $R_{1,1}$ | 1 | If $(a_2 = 2)$ & $(a_5 = 5)$, then $d = 1$ | $p_1, p_{10}$ | 33.33 | 100 |
| 2 | $R_{2,1}$ | 1 | If $(a_5 = 6)$, then $d = 1$ | $p_3, p_8$ | 33.33 | 100 |
| 3 | $R_{3,1}$ | 1 | If $(a_3 = 3)$, then $d = 1$ | $p_9$ | 16.17 | 100 |
| 4 | $R_{4,1}$ | 1 | If $(a_1 = 3)$ & $(a_3 = 1)$, then $d = 1$ | $p_6$ | 16.17 | 100 |
| 5 | $R_{5,2}$ | 2 | If $(a_1 = 1)$ & $(a_5 = 4)$, then $d = 2$ | $p_4, p_7$ | 50 | 100 |
| 6 | $R_{6,2}$ | 2 | If $(a_3 = 1)$ & $(a_4 = 3)$, then $d = 2$ | $p_2$ | 25 | 100 |
| 7 | $R_{7,2}$ | 2 | If $(a_2 = 1)$ & $(a_4 = 1)$, then $d = 2$ | $p_5$ | 25 | 100 |

## 3. FORMAL CONCEPT ANALYSIS AND BACKGROUND

Formal concept analysis (FCA) introduced by R. Wille [22] provides conceptual tools for the analysis of data, and it has applied to many quite different fields such as psychology, sociology, anthropology, medicine, biology, linguistics, computer sciences and industrial engineering. The main purpose of the method is to visualize the data in form of concept lattices and thereby to make them more transparent and more easily discussible and criticisable. At the same time it allows knowledge acquisition from (or by) an expert by putting very precise questions, which either have to be confirmed or to be refuted by a counterexample.

The main objective is to support the user in analyzing and structuring a domain of interest based on mathematization of the concept and conceptual hierarchy. It activates mathematical thinking for conceptual data analysis and knowledge processing based on a formal understanding of a concept as a unit of thought. Concepts can be philosophically understood as the basic units of thought formed in dynamic processes within social and cultural environments. Therefore, by philosophical tradition a concept is comprised of extension and intension. The extension of a formal concept is formed by all objects to which the concept applies and the intension that consists of all attributes existing in those objects. The set of objects, attributes and the relations between an object and an attribute in a dataset form the basic conceptual structure of FCA known as formal context. Concepts can only lay on relationships with many other concepts where the subconcept-superconcept relation plays a vital role. A subconcept of a superconcept means that the extension of the subconcept is contained in the extension of the super concept. This is equivalent to the relationship that the intension of the subconcept contains the intension of the superconcept [23].

## 3.1. Formal Context and Formal Concept

In this section we recall the basic definitions and notations of formal concept analysis developed by R. Wille [22]. A formal context is defined as a set structure $K=(U,A,R)$ consists of two sets $U$ and $A$ while $R$ is a binary relation between $U$ and $A$, *i.e.* $R \subseteq U \times A$ . The elements of $U$ are called the objects and the elements of $A$ are called the attributes of the context. The formal concept of the formal context $(U,A,R)$ is defined with the help of derivation operators. The derivation operators are defined for arbitrary $X \subseteq U$ and $Y \subseteq A$ as follows:

$$X' = \{a \in A : uRa \quad \forall\ u \in X\} \tag{7}$$

$$Y' = \{u \in U : uRa \quad \forall\ a \in A\} \tag{8}$$

A formal concept of a formal context $K=(U,A,R)$ is defined as a pair $(X, Y)$ with $X \subseteq U$ , $Y \subseteq A$, $X = Y'$, and $Y = X'$ . The first member *X*, of the pair $(X, Y)$ is called the extent whereas the second member *Y* is called the intent of the formal concept. Objects in *X* share all properties *Y*, and only properties *Y* are possessed by all objects in *X*. A basic result is that the formal concepts of a formal context are always forming the mathematical structure of a lattice with respect to the subconcept-superconcept relation. Therefore, the set of all formal concepts forms a complete lattice called a concept lattice [22]. The subconcept-superconcept relation can be depicted best by a lattice diagram and we can derive concepts, implication sets, and association rules based on the cross table. In such a diagram the name of each object is attached to its represented object concept and the name of each attribute is attached to its represented attribute concept. The subconcept-superconcept relation is transitive. It means that, a concept is the subconcept of any concept that can be reached by travelling upwards from it. Now we present the cross table of the decision rules for decision class 1 obtained in Table 3 is given in Table 4, where the rows are represented as objects and columns are represented as attributes. The relation between them is represented by a cross.

Table 4. Cross table of decision rules for decision class 1

| Rule Number | Notation | $a_1$ | | | $a_2$ | | | $a_3$ | | | $a_5$ | | |
|---|---|---|---|---|---|---|---|---|---|---|---|---|---|
| | | $A_{11}$ | $A_{12}$ | $A_{13}$ | $A_{21}$ | $A_{22}$ | $A_{23}$ | $A_{31}$ | $A_{32}$ | $A_{33}$ | $A_{54}$ | $A_{55}$ | $A_{56}$ |
| 1 | $R_{1,1}$ | | | | | × | | | | | | × | |
| 2 | $R_{2,1}$ | | | | | | | | | | | | × |
| 3 | $R_{3,1}$ | | | | | | | | | × | | | |
| 4 | $R_{4,1}$ | | | × | | | | × | | | | | |

The corresponding lattice diagram is presented in Figure 1, where the nodes represent formal concepts. In particular, one concept consists of object 1 and attributes $A_{22}$, $A_{55}$, is expressed as ({1}, { $A_{22}, A_{55}$ }). Similarly, another concept consists of object 2 and attribute $A_{56}$ is expressed as ({2}, { $A_{56}$ }). It is also observed that, lines up give more general concepts whereas lines down give more specific concepts. A pair of set of objects and a set of attributes that is closed in this manner is called a formal concept [24].

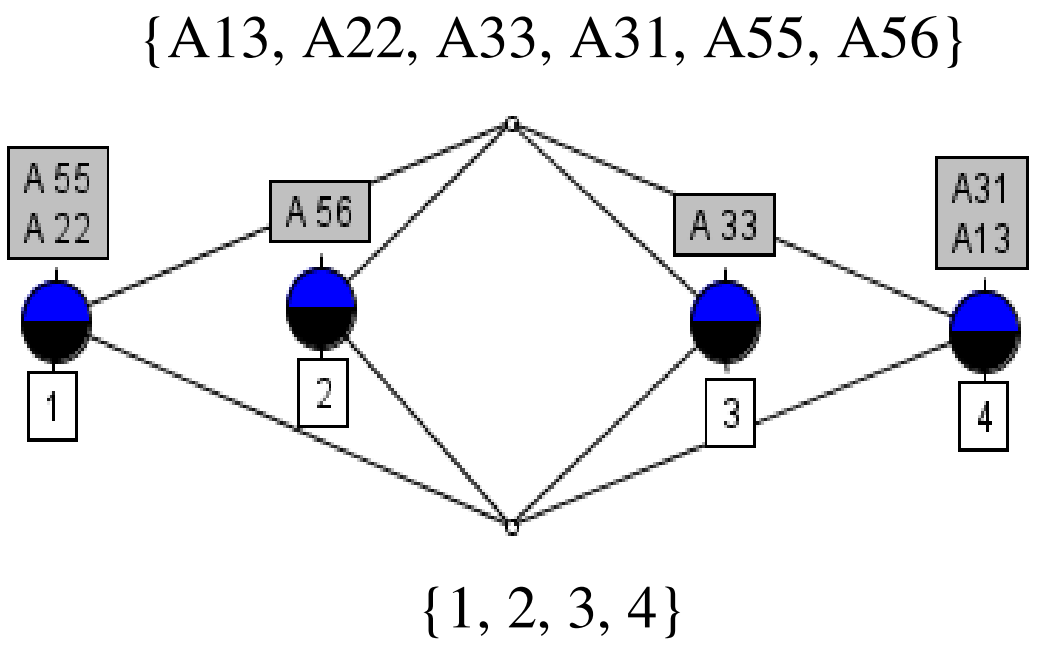

Figure 1. The lattice diagram of the decision rules for decision class 1

## 4. Proposed Intelligent Mining Model

In this section, we propose our intelligent mining model that consists of problem definition, target data, preprocessed data, processed data, data partition, rule discovery, and formal concept analysis as shown in Figure 2.

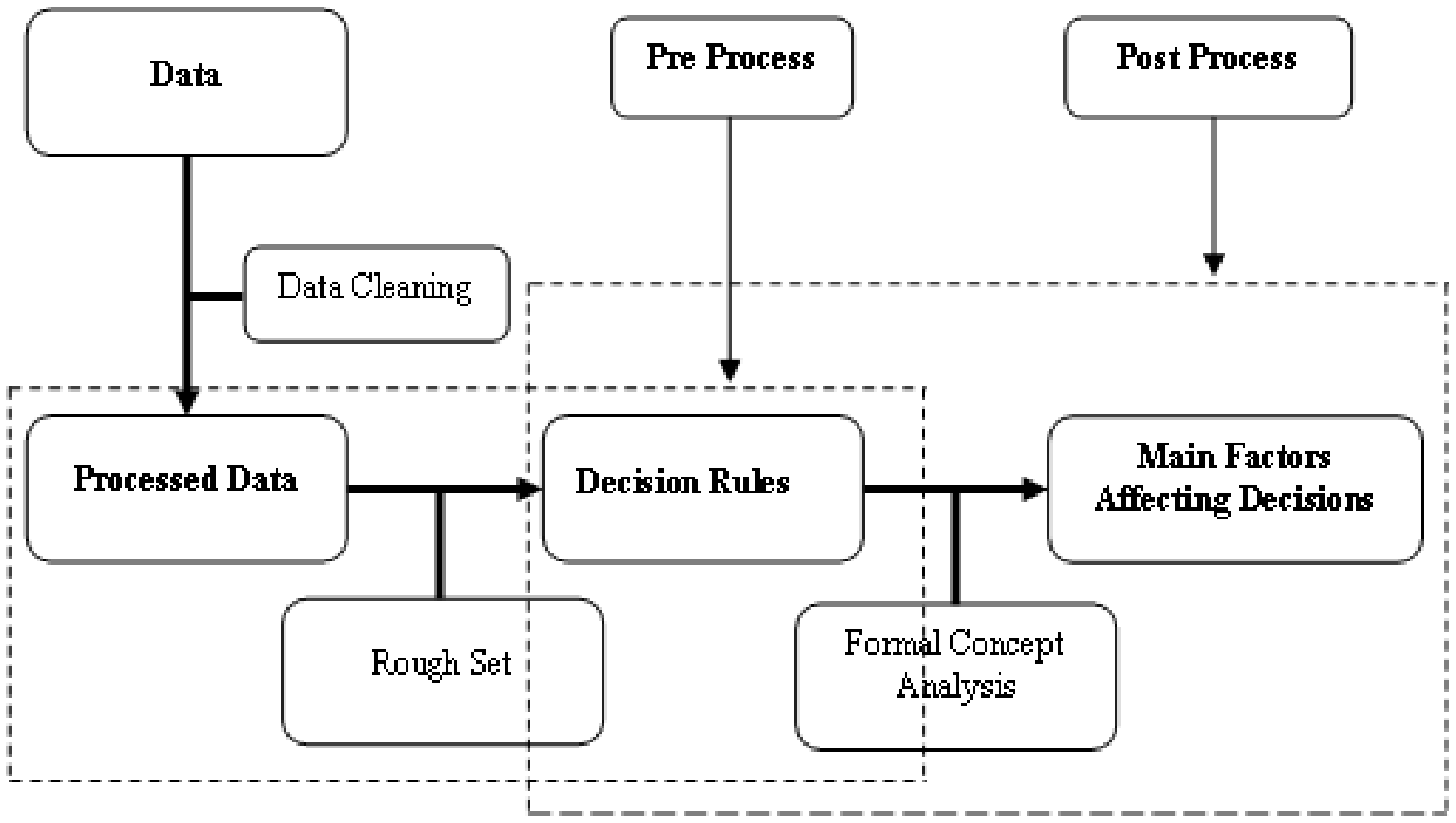

Figure 2. Proposed intelligent mining model

The fundamental step of any model in which we identify the right problem is the problem definition. Incorporation of prior knowledge is always associated with the problem definition however, the potential validity or usefulness of an individual data element or pattern of data elements may change dramatically from individual to individual, organization to organization, or task to task because of the acquisition of knowledge and reasoning may involve in vagueness and incompleteness. It is very difficult for human to find useful information that is hidden in the accumulated voluminous data. Therefore, most important challenge is to retrive data pattern from the accumulated voluminous data. There is much need for dealing with the incomplete and vague information in classification, concept formulation, and data analysis. To this end here we use two processes such as pre process and post process to mine suitable rules and to explore the relationship among the attributes. In pre process as shown in Figure 2 we use rough set theory for processing data, data classification after removal of noise and missing data to mine suitable rules. In post process we use formal concept analysis from these suitable rules to explore better knowledge for decision making.

The motivation behind this study is that the two theories aim at different goals and summarize different types of knowledge. Rough set theory is used for prediction whereas formal concept analysis is used for description. Therefore, the combination of both with domain intelligence leads to better knowledge.

## 4.1. Pre process Architecture Design

In this section, we present our pre process architecture design that consists of problem definition, data preparation, data partition, rule generation, domain intelligence and rule validation as shown in Figure 3. Problem definition and incorporation of prior knowledge are the fundamental steps of any model in which we identify the right problem. Secondly, proper structuring the corresponding objectives and the associated attributes is done. Finally, a target dataset is created on which data mining is to be performed. Before further analysis, a sequence of data cleaning tasks such as consistency check, removing noise and data completeness is done to ensure that the data are as accurate as possible. Now, we discuss in detail the subsequent steps of the pre process architecture design of the proposed model.

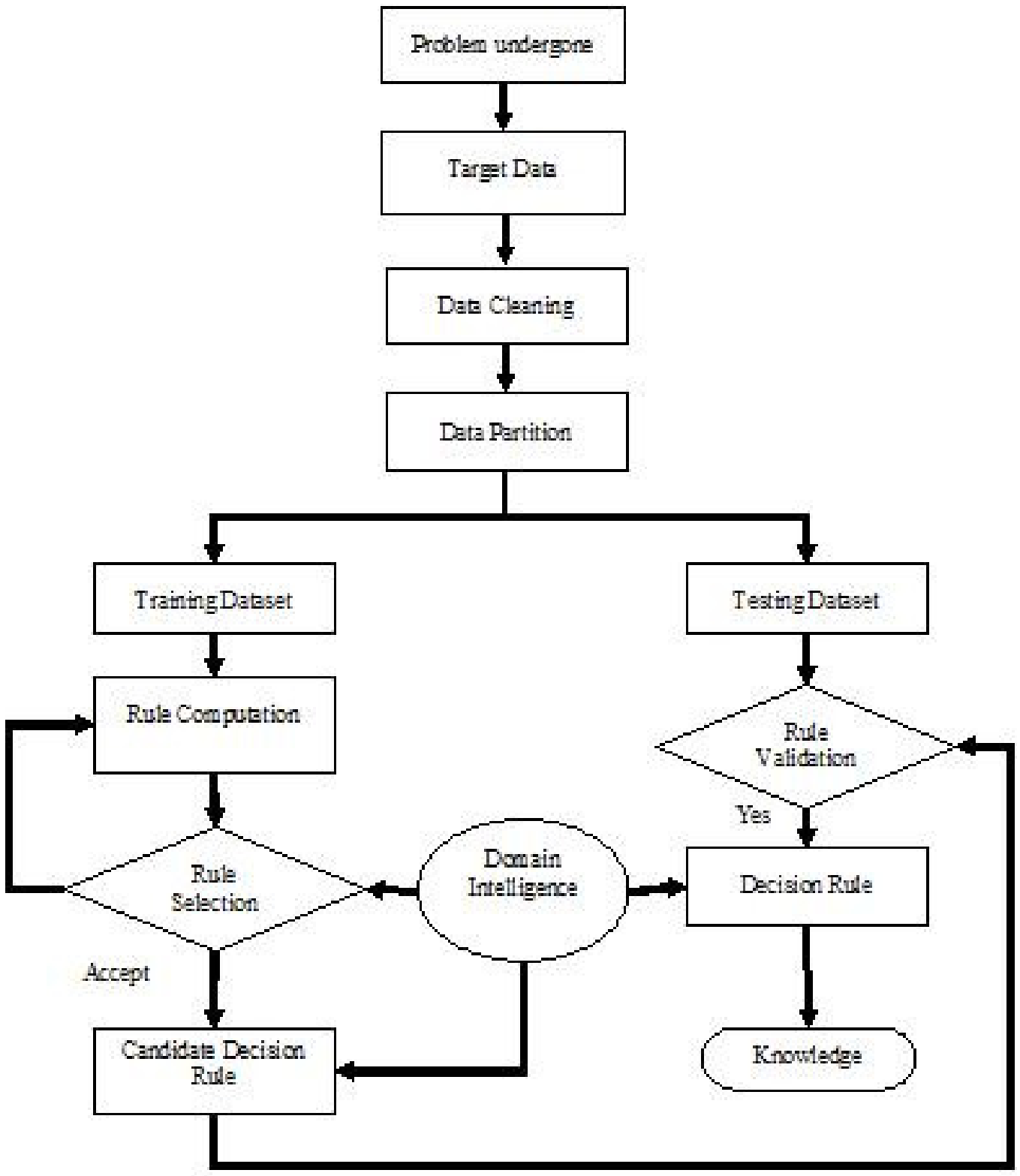

Figure 3. Pre process architecture design

### 4.1.1. Rule Computation Procedure

Now, we propose a rule generation algorithm to generate all the possible reducts by eliminating all dispensable attributes and derive the candidate decision rules from the training dataset as below:

1. Set object number $i = 1$

2. Choose object $i$ from the training data set and compute a set of reducts for all the condition attributes.
3. Replace $i = i + 1$. If all objects have been chosen, then go to step 4; else go to step 2.
4. Compute the number of supporting objects for each reduct after combining the identical reducts.
5. Using the domain intelligent system, evaluate the generated reducts based and compute the number of supporting objects. If candidate decision rules are satisfied, then go to next step 6; else restore the objects associated with the unsatisfied rule and move to step 2.
6. Obtain the decision rules from the selected reducts.
7. Terminate the process and further validate the output results.

#### 4.1.2 Rule-validation Procedure

To examine the objects in the testing dataset to estimate the validity of the rules derived from the rule composing algorithm, we apply the following steps in rule validation procedure.

1. Compare each decision rule obtained from the above algorithm with each new object from the testing data set. Compute the number of objects that support the rule.
2. Repeat step 1 for all decision rules obtained from the rule composing algorithm.
3. Calculate the accuracy of each rule by using the following equation (9).

$$Accuracy = \frac{\text{Total number of supported objects}}{\text{Sum of supported and non-supported objects}} \tag{9}$$

4. If the accuracy is greater than the predefined threshold, then go to step 5; else remove the rule.
5. Process terminates and writes the validated rules.

## 5. An Empirical Study on Heart Diseases

In this empirical study we apply rough set theory for mining knowledge over the data set obtained from different research centers and finally we will find the major factor affecting the decision using formal concept analysis. However, we keep the identity confidential due to some specified official reason. In this study, we have collected historical data from different health research institutions. The 1487 patient's cases were checked for completeness and consistency. In order to avoid unnecessary complexity, we removed unrelated items in the data set. We have removed 553 patient details from that data set as they have no heart disease. In addition to this, 261 patient's data were also removed from the dataset, because none of them had sufficient support. Also, 117 patient's data were removed from the dataset, because of missing attribute values. In total, 931 patient's data were removed from the dataset. Also, we have discussed and attended several meeting with domain experts for getting the clear idea about the heart diseases. This process gave us an understanding of the historical data and essential attribute (symptom) knowledge about the heart disease. However, for completeness of the paper we state briefly about these symptoms.

Table 5. Symptoms notation representation table

| Symptoms | Abbreviation | Notation | Symptoms | Abbreviation | Notation |
|---|---|---|---|---|---|
| Chest pain | CP | $a_1$ | Exercise | EX | $a_7$ |
| Blood pressure | BP | $a_2$ | Old peak | OP | $a_8$ |
| Cholesterol | CH | $a_3$ | Thallium scan | TS | $a_9$ |
| Blood sugar | BS | $a_4$ | Sex | SX | $a_{10}$ |
| Electrocardiography | ECG | $a_5$ | Age | AG | $a_{11}$ |
| Maximum heart rate | MHR | $a_6$ | Type of diagnosis | TD | $d$ |

The most common symptom of heart disease is chest pain and it is of four types, viz. typical angina, atypical angina, non-anginal pain and asymptomatic. The other symptoms included are blood pressure, cholesterol, blood sugar, electrocardiography, maximum heart rate, exercise, old peak, thallium scan, sex and age. Each patient's treatment is different and depends on several factors. Thus, it is essential to identify certain rules and the chief factors so that a patient can identify the disease at an early stage. It can also reduce the financial burden of a patient. Here the data is collected only to diagnosis the heart disease and then suggest the type of heart surgery required. Also, we analyze the historical data to provide the chief factors for each case of the diagnosis. Numerical and literature values based on different symptoms were collected and these parameters became our set of attributes. We consider the diagnosis decision of the patients as the decision variable. The attributes that play major role in heart disease and the notations that are used in our analysis is presented in Table 5. In particular to start with pre process, we randomly divide the 556 patients into the training data set that contained 306 patients (55%) and the testing data set that contained 250 patients (45%). The normalized information is given in Table 6. To make our analysis simpler we have assigned some value to each classification group and these are optional for analysis.

Table 6. Normalized information table

| **Attribute** | **Normalized value** | **Classification** |
|---|---|---|
| Chest pain (CP) | --- | Typical angina (1)<br>Atypical angina (2)<br>Non-anginal pain (3)<br>Asymptomatic (4) |
| Blood pressure (BP) | 120 to 139 / 80 to 90<br>140 to 159 / 90 to 99<br>160 to 179 / 100 to 109<br>≥ 180 / 110 | Normal (1)<br>Medium (2)<br>High (3)<br>Very high (4) |
| Cholesterol (CH)-LDL | <160<br>160-190<br>190-250<br>>250 | Low (1)<br>Medium (2)<br>High (3)<br>Very high (4) |
| Fasting Blood sugar (FBS) | <126<br>>126 | Normal (1)<br>Very high (2) |
| Electrocardiography (ECG) | [-0.5,0.4]<br>[2.45,1.8]<br>[1.4,2.5] | Normal (1)<br>ST-T abnormal (2)<br>Hypertrophy (3) |
| Maximum heart rate (MHR) | <60<br>60-100<br>>100 | Medium (1)<br>Normal (2)<br>High (3) |
| Exercise (EX) | --- | False (1)<br>True (2) |
| Old peak (OP) | <2<br>2-3<br>3> | Low (1)<br>Risk (2)<br>Terrible (3) |
| Thallium scan (TS) | 3<br>6<br>7 | Normal (1)<br>Fixed defect (2)<br>Reversible Defect (3) |
| Sex (SX) | --- | Male (1)<br>Female (2) |
| Age (Age) | <35<br>36-60<br>60-75<br>>75 | Young (1)<br>Mild (2)<br>Old (3)<br>Very old (4) |
| Type of diagnosis (TD) | --- | Hypertensive heart disease (1)<br>Coronary heart disease (2)<br>Heart failure (3)<br>Potential patient (4)<br>Cardiomyopathy (5) |

### 5.1. Pre process of Empirical Study

In this section, we discuss in detail the subsequent steps of the pre process architecture design of the empirical study taken under consideration. The training data set of 306 patients obtained is further classified into five decision classes- decision class 1, 2, 3, 4, and 5- which expresses different diagnosis types: hypertensive heart disease, coronary heart disease, heart failure, potential patient, and cardiomyopathy respectively. The number of patients under each decision class is obtained as 71, 60, 49, 69, and 57 respectively. Here, decision class 1 is selected as an example in this section.

#### 5.1.1. Rule Generation and Selection

We employed the training dataset to derive the reducts and select the final rules with the help of domain intelligence experts. We also removed the unusual outcomes. In addition to this, the identical objects in the training dataset were reduced to only one case in order to avoid unnecessary analysis. Based on the rule generation algorithm, the reducts were determined and presented in Table7. We have generated and summarized the candidate rules based on the number of supporting patients and domain intelligent experts. This is presented in Table 8 for further validation.

Table 7. Rules generated for the decision class 1

| Rules | $a_1$ | $a_2$ | $a_3$ | $a_4$ | $a_5$ | $a_6$ | $a_7$ | $a_8$ | $a_9$ | $a_{10}$ | $a_{11}$ | $d$ | Number of supporting patients |
|---|---|---|---|---|---|---|---|---|---|---|---|---|---|
| [1] | × | 4 | × | × | 2 | × | × | 2 | × | 1 | × | 1 | 8 |
| [2] | 2 | 3 | × | × | × | 3 | 1 | × | × | × | × | 1 | 2 |
| [3] | 3 | 1 | 1 | × | 3 | × | × | × | × | × | × | 1 | 6 |
| [4] | × | × | × | × | 1 | × | 2 | × | × | × | 4 | 1 | 3 |
| [5] | 2 | × | 1 | × | × | × | × | × | 2 | × | × | 1 | 3 |
| [6] | × | × | × | 2 | × | × | × | × | × | × | × | 1 | 2 |
| [7] | × | 4 | 4 | × | × | × | × | × | 3 | × | 2 | 1 | 6 |
| [8] | 3 | × | × | × | × | × | × | 1 | × | × | × | 1 | 3 |
| [9] | × | 1 | 4 | × | × | × | 1 | 3 | × | × | × | 1 | 2 |
| [10] | 2 | 3 | × | × | × | × | × | × | × | × | 1 | 1 | 3 |
| [11] | × | × | 2 | × | × | × | × | × | 1 | 2 | × | 1 | 2 |
| [12] | 1 | × | × | × | 1 | 3 | × | × | × | × | × | 1 | 7 |
| [13] | 3 | 3 | × | × | × | 1 | × | × | × | × | 3 | 1 | 10 |
| [14] | 2 | × | 2 | × | 2 | × | × | × | × | × | × | 1 | 4 |
| [15] | × | × | × | × | 1 | 2 | × | × | × | × | × | 1 | 2 |
| [16] | × | × | × | × | × | 2 | 2 | × | × | × | 1 | 1 | 4 |
| [17] | × | 1 | 3 | × | × | × | 1 | 3 | × | × | × | 1 | 2 |
| [18] | × | 4 | 4 | × | × | × | 1 | 3 | × | × | × | 1 | 2 |

Table 8. candidate rules for hypertensive heart disease

| Rules selected | $a_1$ | $a_2$ | $a_3$ | $a_4$ | $a_5$ | $a_6$ | $a_7$ | $a_8$ | $a_9$ | $a_{10}$ | $a_{11}$ | $d$ | Number of supporting patients |
|---|---|---|---|---|---|---|---|---|---|---|---|---|---|
| [1] | × | 4 | × | × | 2 | × | × | 2 | × | 1 | × | 1 | 8 |
| [2] | 2 | 3 | × | × | × | 3 | 1 | × | × | × | × | 1 | 2 |
| [3] | 3 | 1 | 1 | × | 3 | × | × | × | × | × | × | 1 | 6 |
| [4] | × | × | × | × | 1 | × | 2 | × | × | × | 4 | 1 | 3 |
| [5] | 2 | × | 1 | × | × | × | × | × | 2 | × | × | 1 | 3 |
| [7] | × | 4 | 4 | × | × | × | × | × | 3 | × | 2 | 1 | 7 |
| [9] | × | 1 | 4 | × | × | × | 1 | 3 | × | × | × | 1 | 2 |
| [10] | 2 | 3 | × | × | × | × | × | × | × | × | 1 | 1 | 3 |
| [12] | 1 | × | × | × | 1 | 3 | × | × | × | × | × | 1 | 7 |
| [13] | 3 | 3 | × | × | × | 1 | × | × | × | × | 3 | 1 | 9 |
| [14] | 2 | × | 2 | × | 2 | × | × | × | × | × | × | 1 | 4 |
| [17] | × | 1 | 3 | × | × | × | 1 | 3 | × | × | × | 1 | 2 |
| [18] | × | 4 | 4 | × | × | × | 1 | 3 | × | × | × | 1 | 2 |

#### 5.1.2 The Rule Validation

We used the testing data set to examine the accuracy of each candidate rule generated, to estimate the corresponding validity. Unlike training data set, the testing data set of 250 patients is classified into five decision classes- decision class 1, 2, 3, 4, and 5- which expressed different diagnosis types: hypertensive heart disease, coronary heart disease, heart failure, potential patient, and cardiomyopathy respectively for further validation. The number of patients under each decision class is obtained as 63, 51, 38, 55, and 43 respectively. Finally, we expressed these candidate rules in Table 9. The total number of supporting, non-supporting patients and the accuracy of each candidate rule are mentioned in the right hand column of the Table 9. Here, decision class 1 is selected as an example in this section.

Since the accuracy of candidate rules 4, and 10 is less than the predefined domain intelligence threshold 60%, these two rules are discarded while validation. Thus the rules selected from pre process are 1, 2, 3, 5, 7, 9, 12, 13, 14, 17, 18 is the input to the post process for identifying the chief factors. For example, rule 1 can be stated as: IF blood pressure is very high, electrocardiography (ECG) is ST-T abnormal, maximum heart rate (MHR) is normal, and sex is male THEN we can infer that the heart disease is hypertensive heart disease. Similarly, the other rules can also be obtained from the given Table 9.

Table 9. Candidate rule validation

| Rule | Description | | Support | Non-support |
|---|---|---|---|---|
| [1] | IF | $a_2 = 4$, $a_5 = 2$, $a_8 = 2$, and $a_{10} = 1$ | 8 | 1 |
| | THEN | we can infer that the heart disease is hypertensive heart disease. ($d = 1$) | Accuracy: 88% | |
| [2] | IF | $a_1 = 2$, $a_2 = 3$, $a_6 = 3$, and $a_7 = 1$ | 3 | 1 |
| | THEN | $d = 1$ | Accuracy: 75% | |
| [3] | IF | $a_1 = 3$, $a_2 = 1$, $a_3 = 1$, and $a_5 = 3$ | 6 | 0 |
| | THEN | $d = 1$ | Accuracy: 100% | |
| [4] | IF | $a_3 = 1$, $a_7 = 2$, and $a_{11} = 4$ | 0 | 2 |
| | THEN | $d = 1$ | Accuracy: 0% | |
| [5] | IF | $a_1 = 2$, $a_3 = 1$, and $a_9 = 2$ | 3 | 1 |
| | THEN | $d = 1$ | Accuracy: 75% | |
| [7] | IF | $a_2 = 4$, $a_3 = 4$, $a_9 = 3$, and $a_{11} = 2$ | 6 | 1 |
| | THEN | $d = 1$ | Accuracy: 85% | |
| [9] | IF | $a_2 = 1$, $a_3 = 4$, $a_7 = 1$, and $a_8 = 3$ | 2 | 1 |
| | THEN | $d = 1$ | Accuracy: 66% | |
| [10] | IF | $a_1 = 2$, $a_2 = 3$, and $a_{11} = 1$ | 1 | 2 |
| | THEN | $d = 1$ | Accuracy: 33% | |
| [12] | IF | $a_1 = 1$, $a_3 = 1$, and $a_6 = 3$ | 6 | 1 |
| | THEN | $d = 1$ | Accuracy: 85% | |
| [13] | IF | $a_1 = 3$, $a_2 = 3$, $a_6 = 1$, and $a_{11} = 3$ | 5 | 1 |
| | THEN | $d = 1$ | Accuracy: 83% | |
| [14] | IF | $a_1 = 2$, $a_3 = 2$, and $a_5 = 2$ | 3 | 2 |
| | THEN | $d = 1$ | Accuracy: 60% | |
| [17] | IF | $a_2 = 1$, $a_3 = 3$, $a_7 = 1$, and $a_8 = 3$ | 2 | 1 |
| | THEN | $d = 1$ | Accuracy: 66% | |
| [18] | IF | $a_2 = 4$, $a_3 = 4$, $a_7 = 1$, and $a_8 = 3$ | 3 | 1 |
| | THEN | $d = 1$ | Accuracy: 75% | |

## 5.2. Post process of Empirical Study

Formal concept analysis can do the data classification. However, data was already classified in the pre process. The objective of this process is to use formal concept analysis to aggregate the suitable rules that are validated from pre process and hence to obtain the chief factors affecting the decisions. This helps the decision maker to identify the type of disease and its chief characteristics at an early stage. We discuss the results and discussions of the decision class 1-hypertensive heart disease in section 5.2.1. In section 5.2.2 we discuss the results of the decision class 2-coronary heart disease. The result of the decision class 3-heart failure is discussed in section 5.2.3. Section 5.2.4 discusses the results of decision class 4-potential patients whereas the results of decision class 5-cardiomyopathy is discussed in section 5.2.5.

### 5.2.1 Decision Class Hypertensive Heart Disease

Now we present the context table in Table 10, which converted 11 rules of hypertensive heart disease representing the attributes obtained from pre process. In Figure 4, we present the lattice diagram of the context Table 10 for the decision class hypertensive heart disease.

Table 10. Context table of decision class hypertensive heart disease

| Rule | $a_1$ | | | | $a_2$ | | | | $a_3$ | | | | $a_5$ | | | $a_6$ | | | $a_7$ | | $a_8$ | | | $a_9$ | | | $a_{10}$ | | $a_{11}$ | | | |
|---|---|---|---|---|---|---|---|---|---|---|---|---|---|---|---|---|---|---|---|---|---|---|---|---|---|---|---|---|---|---|---|---|
| | $a_{11}$ | $a_{12}$ | $a_{13}$ | $a_{14}$ | $a_{21}$ | $a_{22}$ | $a_{23}$ | $a_{24}$ | $a_{31}$ | $a_{32}$ | $a_{33}$ | $a_{34}$ | $a_{51}$ | $a_{52}$ | $a_{53}$ | $a_{61}$ | $a_{62}$ | $a_{63}$ | $a_{71}$ | $a_{72}$ | $a_{81}$ | $a_{82}$ | $a_{83}$ | $a_{91}$ | $a_{92}$ | $a_{93}$ | $a_{10,1}$ | $a_{10,2}$ | $a_{11,1}$ | $a_{11,2}$ | $a_{11,3}$ | $a_{11,4}$ |
| [1] | | | | | | | | x | | | | | | x | | | | | | | | x | | | | | x | | | | | |
| [2] | | x | | | | | x | | | | | | | | | | | x | x | | | | | | | | | | | | | |
| [3] | | | x | | x | | | | x | | | | | | x | | | | | | | | | | | | | | | | | |
| [5] | | x | | | | | | | x | | | | | | | | | | | | | | | | x | | | | | | | |
| [7] | | | | | | | | x | | | | x | | | | | | | | | | | | | | x | | | | x | | |
| [9] | | | | | x | | | | | | | x | | | | | | | x | | | | x | | | | | | | | | |
| [12] | x | | | | | | | | | | | | x | | | | | x | | | | | | | | | | | | | | |
| [13] | | | x | | | | x | | | | | | | | | | x | | | | | | | | | | | | | | x | |
| [14] | | x | | | | | | | | x | | | | x | | | | | | | | | | | | | | | | | | |
| [17] | | | | | x | | | | | | x | | | | | | | | x | | | | x | | | | | | | | | |
| [18] | | | | | | | | x | | | | x | | | | | | | x | | | | x | | | | | | | | | |

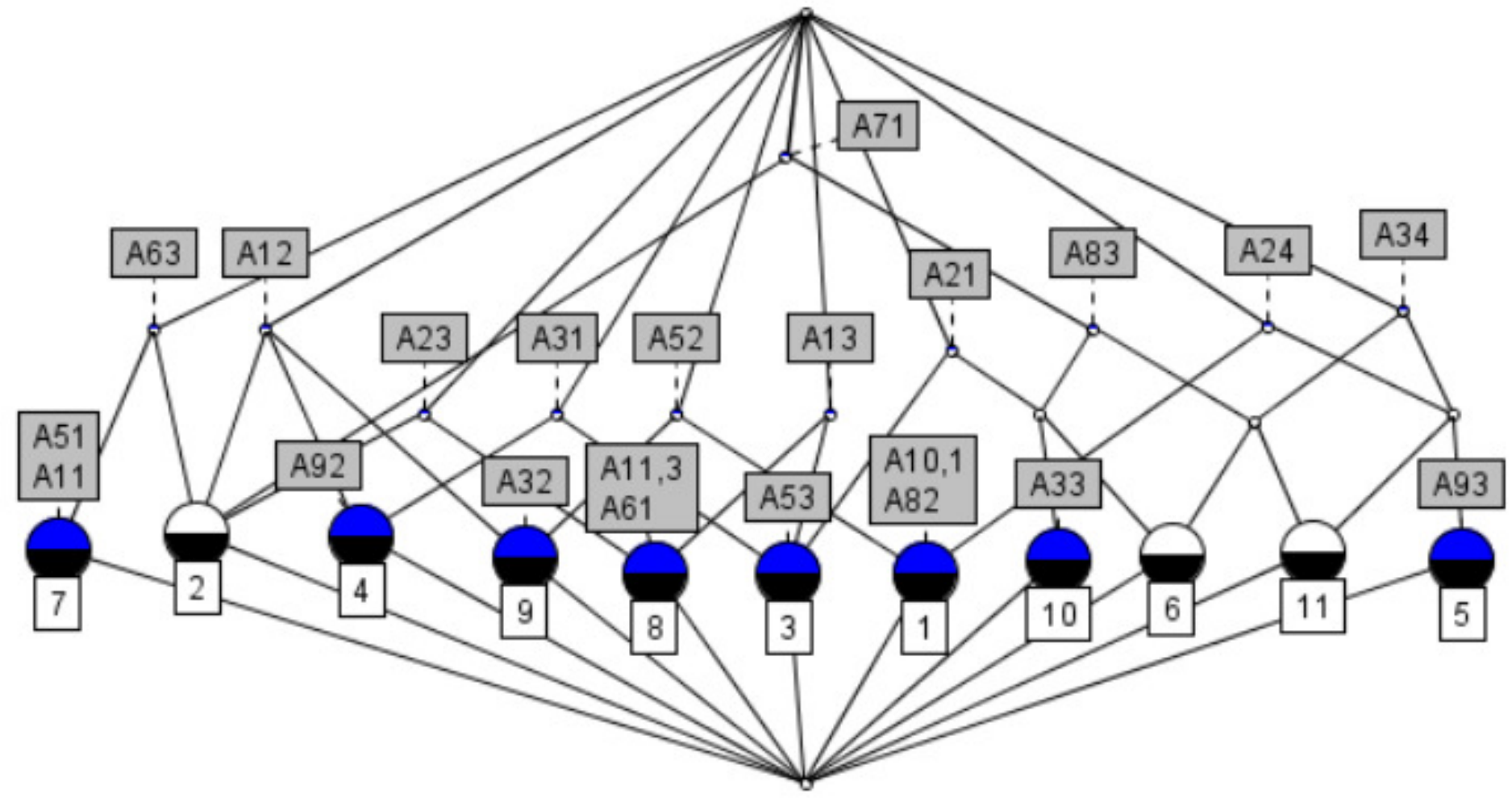

Figure 4 Lattice diagram of the decision class hypertensive heart disease

The implication set table for the decision class hypertensive heart disease is presented in Table 11. Further we compute the implication relation table from the implication set Table 11 and is given in Table 12. From the higher frequency of the implication Table 12, we can find the chief characteristics influencing the decision class hypertensive heart disease. The most important characteristics of this class is old peak-terrible. The next important characteristic that leads to this class is exercise-false. However, the other characteristics that are equally important for this class of disease are chest pain-atypical angina, blood pressure-high, and MHR-high.

Table 11. Implication set table for decision class hypertensive heart disease

| **Implication Sets** | | |
|---|---|---|
| 1 < 3 > A83 ==> A71; | 11 < 1 > A32 ==> A12 A52; | 21 < 1 > A12 A71 ==> A23 A63; |
| 2 < 2 > A21 A71 ==> A83; | 12 < 1 > A33 ==> A21 A71 A83; | 22 < 1 > A23 A71 ==> A12 A63; |
| 3< 2 > A34 A71 ==> A83; | 13 < 1 > A21 A34 ==> A71 A83; | 23 < 1 > A24 A71 ==> A34 A83; |
| 4 < 1 > A11 ==> A51 A63; | 14 < 1 > A51 ==> A11 A63; | 24 < 1 > A63 A71 ==> A12 A23; |
| 5 < 1 > A13 A21 ==> A31 A53; | 15 < 1 > A12 A52 ==> A32; | 25 < 1 > A82 ==> A24 A52 A101; |
| 6 < 1 > A12 A23 ==> A63 A71; | 16 < 1 > A24 A52 ==> A82 A101; | 26 < 1 > A92 ==> A12 A31; |
| 7 < 1 > A13 A23 ==> A61 A113; | 17 < 1 > A53 ==> A13 A21 A31; | 27 < 1 > A93 ==> A24 A34; |
| 8 < 1 > A12 A31 ==> A92; | 18 < 1 > A61 ==> A13 A23 A113; | 28 < 1 > A101 ==> A24 A52 A82; |
| 9 < 1 > A13 A31 ==> A21 A53; | 19 < 1 > A12 A63 ==> A23 A71; | 29 < 1 > A113 ==> A13 A23 A61; |
| 10 < 1 > A21 A31 ==> A13 A53; | 20 < 1 > A23 A63 ==> A12 A71; | |

Table 12. Implication relation table for decision class hypertensive heart disease

| **Superconcept** | $a_{11}$ | $a_{12}$ | $a_{13}$ | $a_{21}$ | $a_{23}$ | $a_{24}$ | $a_{31}$ | $a_{32}$ | $a_{34}$ | $a_{51}$ |
|---|---|---|---|---|---|---|---|---|---|---|
| **Subconcept** | $a_{51}$ | $a_{32}$<br>$a_{23}, a_{63}$<br>$a_{23}, a_{71}$<br>$a_{63}, a_{71}$<br>$a_{92}$ | $a_{21}, a_{31}$<br>$a_{53}$<br>$a_{61}$<br>$a_{11,3}$ | $a_{13}, a_{31}$<br>$a_{33}$<br>$a_{53}$ | $a_{61}$<br>$a_{12}, a_{63}$<br>$a_{12}, a_{71}$<br>$a_{63}, a_{71}$<br>$a_{11,3}$ | $a_{82}$<br>$a_{93}$<br>$a_{10,1}$ | $a_{13}, a_{21}$<br>$a_{53}$<br>$a_{92}$ | $a_{12}, a_{52}$ | $a_{24}, a_{71}$<br>$a_{93}$ | $a_{11}$ |
| **Frequency** | 1 | 8 | 5 | 4 | 8 | 3 | 4 | 2 | 3 | 1 |
| **Superconcept** | $a_{52}$ | $a_{53}$ | $a_{61}$ | $a_{63}$ | $a_{71}$ | $a_{82}$ | $a_{83}$ | $a_{92}$ | $a_{10,1}$ | $a_{11,3}$ |
| **Subconcept** | $a_{32}$<br>$a_{82}$<br>$a_{10,1}$ | $a_{13}, a_{21}$<br>$a_{13}, a_{31}$<br>$a_{21}, a_{31}$ | $a_{13}, a_{23}$<br>$a_{11,3}$ | $a_{11}$<br>$a_{12}, a_{23}$<br>$a_{51}$<br>$a_{12}, a_{71}$<br>$a_{23}, a_{71}$ | $a_{83}$ *3<br>$a_{12}, a_{23}$<br>$a_{33}$<br>$a_{21}, a_{34}$<br>$a_{12}, a_{63}$<br>$a_{23}, a_{63}$ | $a_{24}, a_{52}$<br>$a_{10,1}$ | $a_{21}, a_{71}$ *2<br>$a_{34}, a_{71}$ *2<br>$a_{33}$<br>$a_{21}, a_{34}$<br>$a_{24}, a_{71}$ | $a_{12}, a_{31}$ | $a_{24}, a_{52}$<br>$a_{82}$ | $a_{13}, a_{23}$<br>$a_{61}$ |
| **Frequency** | 3 | 6 | 3 | 8 | 12 | 3 | 13 | 2 | 3 | 3 |

### 5.2.2 Decision Class Coronary Heart Disease

The number of rules generated using rule generation algorithm due to rough set theory is 19 and is further minimized to 15 by using domain intelligence experts. This is further validated and minimized to 13 numbers of rules using testing data set. Now we present the context table in Table 13, which converted 13 rules of coronary heart disease representing the attributes obtained from pre process. In Figure 5, we present the lattice diagram of the context Table 13 for the decision class coronary heart disease. The implication set table is presented in Table 14 whereas the implication relation table is presented in Table 15.

Table 13. Context table of decision class coronary heart disease

| **Rule** | $a_1$ | | | | $a_2$ | | | | $a_3$ | | | | $a_4$ | | $a_5$ | | | $a_6$ | | | $a_7$ | | $a_8$ | | | $a_9$ | | | $a_{10}$ | | $a_{11}$ | | | |
|---|---|---|---|---|---|---|---|---|---|---|---|---|---|---|---|---|---|---|---|---|---|---|---|---|---|---|---|---|---|---|---|---|---|---|
| | $a_{11}$ | $a_{12}$ | $a_{13}$ | $a_{14}$ | $a_{21}$ | $a_{22}$ | $a_{23}$ | $a_{24}$ | $a_{31}$ | $a_{32}$ | $a_{33}$ | $a_{34}$ | $a_{41}$ | $a_{42}$ | $a_{51}$ | $a_{52}$ | $a_{53}$ | $a_{61}$ | $a_{62}$ | $a_{63}$ | $a_{71}$ | $a_{72}$ | $a_{81}$ | $a_{82}$ | $a_{83}$ | $a_{91}$ | $a_{92}$ | $a_{93}$ | $a_{10,1}$ | $a_{10,2}$ | $a_{11,1}$ | $a_{11,2}$ | $a_{11,3}$ | $a_{11,4}$ |
| [1] | | | | | | | | | × | | | | | | | | | × | | | | | | × | | | | × | | | | | | |
| [2] | | | | | | × | | | | | | × | | | | | | | | | | | × | | | | | | | | | | | |
| [3] | | × | | | | | | × | | | | | | | | | | × | | | × | | | | | | × | | | | | | | |
| [4] | | | × | | | | | | | | | | | | | | × | × | | | | | | | | | × | | | | | | | |
| [5] | | | | | | | | × | | | | | | | | × | | | | | × | | | | | | | | × | | × | | | |
| [6] | | × | | | | | | | | | × | | | | | | | | | | | | | | | | | × | | × | | | × | |
| [7] | | | × | | | | × | | | | | | | | | | | | | | | | | | × | | | | | | | | × | |
| [8] | | × | | | | × | | | | | | | | | | | | | | | | | | | | | | × | | | | | | × |
| [9] | | | | | | | × | | | | | | | | | | | | | | × | | | | | | | × | | × | | | | × |
| [10] | | | | | | | | × | | | | × | | | | | × | | | | | | | | | | | | | × | | | | |
| [11] | | | | | | | | | × | | | | | | | | × | | | | | | | | | | | | | | × | | | |
| [12] | | | | | | | | | | | | | × | | | | | × | | | | | | | | | | × | | | | | | |
| [13] | | | | × | | | | × | | | | × | | | | | | | | | | | | | | | | | | | | | | × |

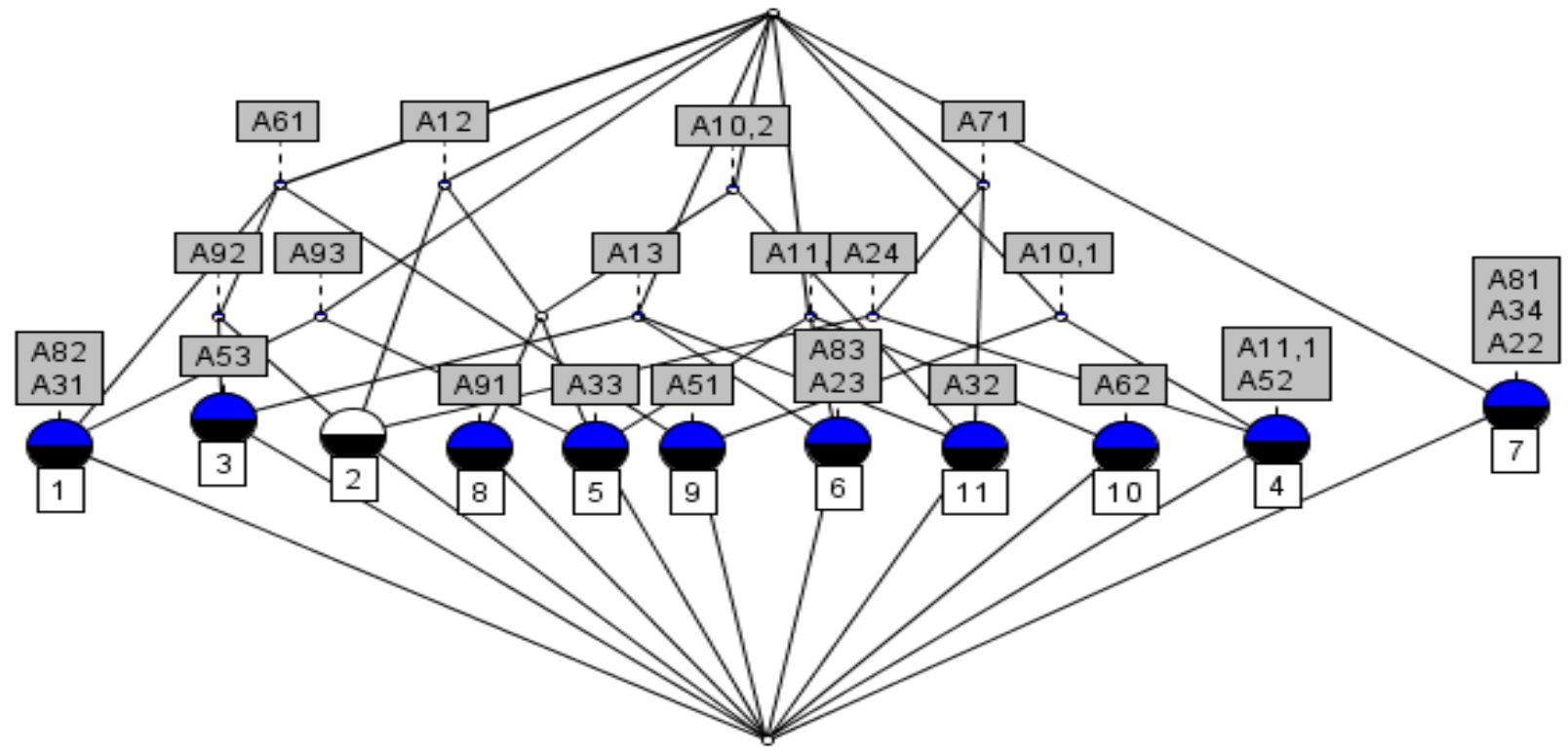

Figure 5 Lattice diagram of the decision class coronary heart disease

Table 14. Implication set table for decision class coronary heart disease

| Implication Sets | | |
|---|---|---|
| 1 < 1 > A13 A23 ==> A83 A113 | 20 < 1 > A83 ==> A13 A23 A113 | 39 < 1 > A92 A113 ==> A12 A33 A102 |
| 2 < 1 > A12 A24 ==> A61 A71 A92 | 21 < 1 > A13 A92 ==> A53 A61 | 40 < 1 > A102 A113 ==> A12 A33 A92 |
| 3 < 1 > A33 ==> A12 A92 A102 A113 | 22 < 1 > A24 A92 ==> A12 A61 A71 | 41 < 1 > A12 A114 ==> A22 A93 |
| 4 < 1 > A22 A34 ==> A81 | 23 < 1 > A53 A92 ==> A13 A61 | 42 < 1 > A22 A114 ==> A12 A93 |
| 5 < 1 > A41 ==> A61 A93 | 24 < 1 > A71 A92 ==> A12 A24 A61 | 43 < 1 > A23 A114 ==> A71 A93 A102 |
| 6 < 1 > A52 ==> A24 A71 A101 A111 | 25 < 1 > A31 A93 ==> A61 A82 | 44 < 1 > A24 A114 ==> A14 A34 |
| 7 < 1 > A13 A53 ==> A61 A92 | 26 < 1 > A101 ==> A24 A52 A71 A111 | 45 < 1 > A34 A114 ==> A14 A24 |
| 8 < 1 > A24 A53 ==> A34 A102 | 27 < 1 > A12 A102 ==> A33 A92 A113 | 46 < 1 > A71 A114 ==> A23 A93 A102 |
| 9 < 1 > A31 A53 ==> A111 | 28 < 1 > A24 A102 ==> A34 A53 | 47 < 1 > A102 A114 ==> A23 A71 A93 |
| 10 < 1 > A34 A53 ==> A24 A102 | 29 < 1 > A34 A102 ==> A24 A53 | 48 < 1 > A14 ==> A24 A34 A114 |
| 11 < 1 > A12 A61 ==> A24 A71 A92 | 30 < 1 > A53 A102 ==> A24 A34 | 49 < 1 > A12 A22 ==> A93 A114 |
| 12 < 1 > A13 A61 ==> A53 A92 | 31 < 1 > A92 A102 ==> A12 A33 A113 | 50 < 1 > A23 A71 ==> A93 A102 A114 |
| 13 < 1 > A24 A61 ==> A12 A71 A92 | 32 < 1 > A24 A111 ==> A52 A71 A101 | 51 < 1 > A12 A93 ==> A22 A114 |
| 14 < 1 > A31 A61 ==> A82 A93 | 33 < 1 > A31 A111 ==> A53 | 52 < 1 > A22 A93 ==> A12 A114 |
| 15 < 1 > A53 A61 ==> A13 A92 | 34 < 1 > A53 A111 ==> A31 | 53 < 1 > A23 A93 ==> A71 A102 A114 |
| 16 < 1 > A12 A71 ==> A24 A61 A92 | 35 < 1 > A71 A111 ==> A24 A52 A101 | 54 < 1 > A71 A93 ==> A23 A102 A114 |
| 17 < 1 > A61 A71 ==> A12 A24 A92 | 36 < 1 > A12 A113 ==> A33 A92 A102 | 55 < 1 > A23 A102 ==> A71 A93 A114 |
| 18 < 1 > A81 ==> A22 A34 | 37 < 1 > A13 A113 ==> A23 A83 | 56 < 1 > A71 A102 ==> A23 A93 A114 |
| 19 < 1 > A82 ==> A31 A61 A93 | 38 < 1 > A23 A113 ==> A13 A83 | 57 < 1 > A93 A102 ==> A23 A71 A114 |

Table 15. Implication relation table for decision class coronary heart disease

| **Superconcept** | $a_{12}$ | $a_{13}$ | $a_{14}$ | $a_{22}$ | $a_{23}$ | $a_{24}$ | $a_{31}$ | $a_{33}$ | $a_{34}$ | $a_{52}$ | $a_{53}$ | $a_{61}$ |
|---|---|---|---|---|---|---|---|---|---|---|---|---|
| **Subconcept** | $a_{33}$<br>$a_{92}, a_{11,3}$<br>$a_{10,2}, a_{11,3}$<br>$a_{22}, a_{11,4}$<br>$a_{22}, a_{93}$<br>$a_{24}, a_{61}$<br>$a_{61}, a_{71}$<br>$a_{24}, a_{92}$<br>$a_{71}, a_{92}$<br>$a_{92}, a_{10,2}$ | $a_{53}, a_{61}$<br>$a_{83}$<br>$a_{53}, a_{92}$<br>$a_{23}, a_{11,3}$ | $a_{24}, a_{11,4}$<br>$a_{34}, a_{11,4}$ | $a_{12}, a_{11,4}$<br>$a_{12}, a_{93}$<br>$a_{81}$ | $a_{93}, a_{10,2}$<br>$a_{71}, a_{10,2}$<br>$a_{71}, a_{93}$<br>$a_{71}, a_{11,4}$<br>$a_{10,2}, a_{11,4}$<br>$a_{83}$<br>$a_{13}, a_{11,3}$ | $a_{52}$<br>$a_{34}, a_{53}$<br>$a_{12}, a_{61}$<br>$a_{12}, a_{71}$<br>$a_{61}, a_{71}$<br>$a_{71}, a_{92}$<br>$a_{10,1}$<br>$a_{34}, a_{10,2}$<br>$a_{53}, a_{10,2}$<br>$a_{71}, a_{11,1}$<br>$a_{34}, a_{11,4}$<br>$a_{14}$ | $a_{53}, a_{11,1}$ | $a_{92}, a_{11,3}$<br>$a_{10,2}, a_{11,3}$<br>$a_{12}, a_{10,2}$<br>$a_{92}, a_{10,2}$<br>$a_{12}, a_{11,3}$ | $a_{24}, a_{11,4}$<br>$a_{14}$<br>$a_{24}, a_{53}$<br>$a_{81}$<br>$a_{24}, a_{10,2}$<br>$a_{53}, a_{10,2}$ | $a_{10,1}$<br>$a_{24}, a_{11,1}$<br>$a_{71}, a_{11,1}$ | $a_{13}, a_{61}$<br>$a_{13}, a_{92}$<br>$a_{24}, a_{10,2}$<br>$a_{31}, a_{11,1}$<br>$a_{34}, a_{10,2}$ | $a_{12}, a_{24}$<br>$a_{41}$<br>$a_{13}, a_{53}$<br>$a_{12}, a_{71}$<br>$a_{82}$<br>$a_{13}, a_{92}$<br>$a_{24}, a_{92}$<br>$a_{53}, a_{92}$<br>$a_{71}, a_{92}$<br>$a_{31}, a_{93}$ |
| **Frequency** | 19 | 7 | 4 | 5 | 13 | 21 | 2 | 10 | 10 | 5 | 10 | 18 |
| **Superconcept** | $a_{71}$ | $a_{81}$ | $a_{82}$ | $a_{83}$ | $a_{92}$ | $a_{93}$ | $a_{10,1}$ | $a_{10,2}$ | $a_{11,1}$ | $a_{11,3}$ | $a_{11,4}$ | |
| **Subconcept** | $a_{12}, a_{24}$<br>$a_{52}$<br>$a_{12}, a_{61}$<br>$a_{24}, a_{61}$<br>$a_{24}, a_{92}$<br>$a_{10,1}$<br>$a_{24}, a_{11,1}$<br>$a_{23}, a_{11,4}$<br>$a_{10,2}, a_{11,4}$<br>$a_{23}, a_{93}$<br>$a_{23}, a_{10,2}$<br>$a_{93}, a_{10,2}$ | $a_{22}, a_{34}$ | $a_{31}, a_{61}$<br>$a_{31}, a_{93}$ | $a_{13}, a_{23}$<br>$a_{13}, a_{11,3}$<br>$a_{23}, a_{11,3}$ | $a_{12}, a_{24}$<br>$a_{13}, a_{53}$<br>$a_{13}, a_{61}$<br>$a_{53}, a_{61}$<br>$a_{12}, a_{10,2}$<br>$a_{12}, a_{11,3}$<br>$a_{10,2}, a_{11,3}$<br>$a_{33}$ | $a_{41}$<br>$a_{31}, a_{61}$<br>$a_{82}$<br>$a_{12}, a_{11,4}$<br>$a_{22}, a_{11,4}$<br>$a_{23}, a_{11,4}$<br>$a_{71}, a_{11,4}$<br>$a_{10,2}, a_{11,4}$<br>$a_{12}, a_{22}$<br>$a_{23}, a_{71}$<br>$a_{23}, a_{10,2}$<br>$a_{71}, a_{10,2}$ | $a_{52}$<br>$a_{24}, a_{11,1}$<br>$a_{71}, a_{11,1}$ | $a_{33}$<br>$a_{24}, a_{53}$<br>$a_{34}, a_{53}$<br>$a_{23}, a_{71}$<br>$a_{23}, a_{93}$<br>$a_{71}, a_{93}$<br>$a_{92}, a_{11,3}$<br>$a_{23}, a_{11,4}$ | $a_{31}, a_{53}$<br>$a_{10,1}$ | $a_{33}$<br>$a_{83}$<br>$a_{13}, a_{23}$<br>$a_{92}, a_{10,2}$<br>$a_{12}, a_{10,2}$ | $a_{14}$<br>$a_{12}, a_{22}$<br>$a_{23}, a_{71}$<br>$a_{12}, a_{93}$<br>$a_{22}, a_{93}$<br>$a_{23}, a_{93}$<br>$a_{71}, a_{93}$<br>$a_{23}, a_{10,2}$<br>$a_{71}, a_{10,2}$<br>$a_{93}, a_{10,2}$ | |
| **Frequency** | 22 | 2 | 4 | 6 | 15 | 22 | 5 | 15 | 3 | 8 | 19 | |

From the higher frequency of the implication Table 15, we find the chief characteristics influencing the decision class coronary heart disease. The most important characteristics of this class are thallium scan-reversible defect and exercise is false. The next important characteristic that leads to this class is blood pressure-very high. However, the other characteristics that are equally important for this class of disease are chest pain-atypical angina and is generally found in very old age.

### 5.2.3 Decision Class Heart Failure

The number of rules generated using rule generation algorithm is 20 and is further minimized to 18 by using domain intelligence experts and further validation. The context table is presented in Table 16, whereas in Figure 6 we present the lattice diagram of the context Table 16. The implication set table for the decision class heart failure is presented in Table 17 whereas the implication relation table is given in Table 18.

Table 16. Context table of decision class heart failure

| Rule | $a_1$ | | | | $a_2$ | | | | $a_3$ | | | | $a_4$ | | $a_5$ | | | $a_6$ | | | $a_7$ | | $a_8$ | | | $a_9$ | | | $a_{10}$ | | $a_{11}$ | | | |
|---|---|---|---|---|---|---|---|---|---|---|---|---|---|---|---|---|---|---|---|---|---|---|---|---|---|---|---|---|---|---|---|---|---|---|
| | $a_{11}$ | $a_{12}$ | $a_{13}$ | $a_{14}$ | $a_{21}$ | $a_{22}$ | $a_{23}$ | $a_{24}$ | $a_{31}$ | $a_{32}$ | $a_{33}$ | $a_{34}$ | $a_{41}$ | $a_{42}$ | $a_{51}$ | $a_{52}$ | $a_{53}$ | $a_{61}$ | $a_{62}$ | $a_{63}$ | $a_{71}$ | $a_{72}$ | $a_{81}$ | $a_{82}$ | $a_{83}$ | $a_{91}$ | $a_{92}$ | $a_{93}$ | $a_{10.1}$ | $a_{10.2}$ | $a_{11.1}$ | $a_{11.2}$ | $a_{11.3}$ | $a_{11.4}$ |
| [1] | | | | | | × | | | | | × | | | | | | | | | | | | | | × | | | | | | | | | |
| [2] | | | × | | | | | | | | | | | | | | | | | × | | × | | | | | | | | | | | × | |
| [3] | | | | | × | | | | | × | | | | | | × | | | | | | | | | | | | | | | | | | |
| [4] | | | | × | | | | × | | | | | | | | | | | | | | | | | | | | | | × | × | | | |
| [5] | | | | | × | | | | | | | | | | × | | | | | | | | | × | | | | × | | | | | | |
| [6] | | | × | | | | | | | | | | | | | | | | | | × | | | | | | × | | | | × | | | |
| [7] | | | | | | | × | | | | | × | | | | × | | | | | | × | | | | | | | | | | | | |
| [8] | | | | | | | | | | | | | | | | × | | | | | | | | | × | | × | | | | | × | | |
| [9] | | | | | | | | × | | | | | | | | | | | | | | | × | | | | | | | | | | | × |
| [10] | | | | | | × | | | | | | | | | | | | | | | | | | | | | | | | × | | | × | |
| [11] | | × | | | | | | | | | | | | | | | | | | | | | | | × | | × | | | | | | × | |
| [12] | | | | | | | | | × | | | | | | | | × | | | | | | | | | | | | | | × | | | |
| [13] | | | | | | | | | | | | | × | | | | | × | | | | | | | | | | × | | | | | | |
| [14] | | | | × | | | | × | | | | × | | | | | | | | | | | | | | | | | | | | | | × |
| [15] | × | | | | | | | | | | | | | | | | | | | | | | | | × | | | | | | | | × | |
| [16] | | | × | | | | | | | | × | | | | | | | × | | | | × | | | | | | | | | | | | |
| [17] | | | | | | | | | | | × | | | | | | | | | | | | × | | | | | | | | | × | | |
| [18] | | | | × | | | | × | | | | | | | | | | | | | | × | | | | | | | | | | | | × |

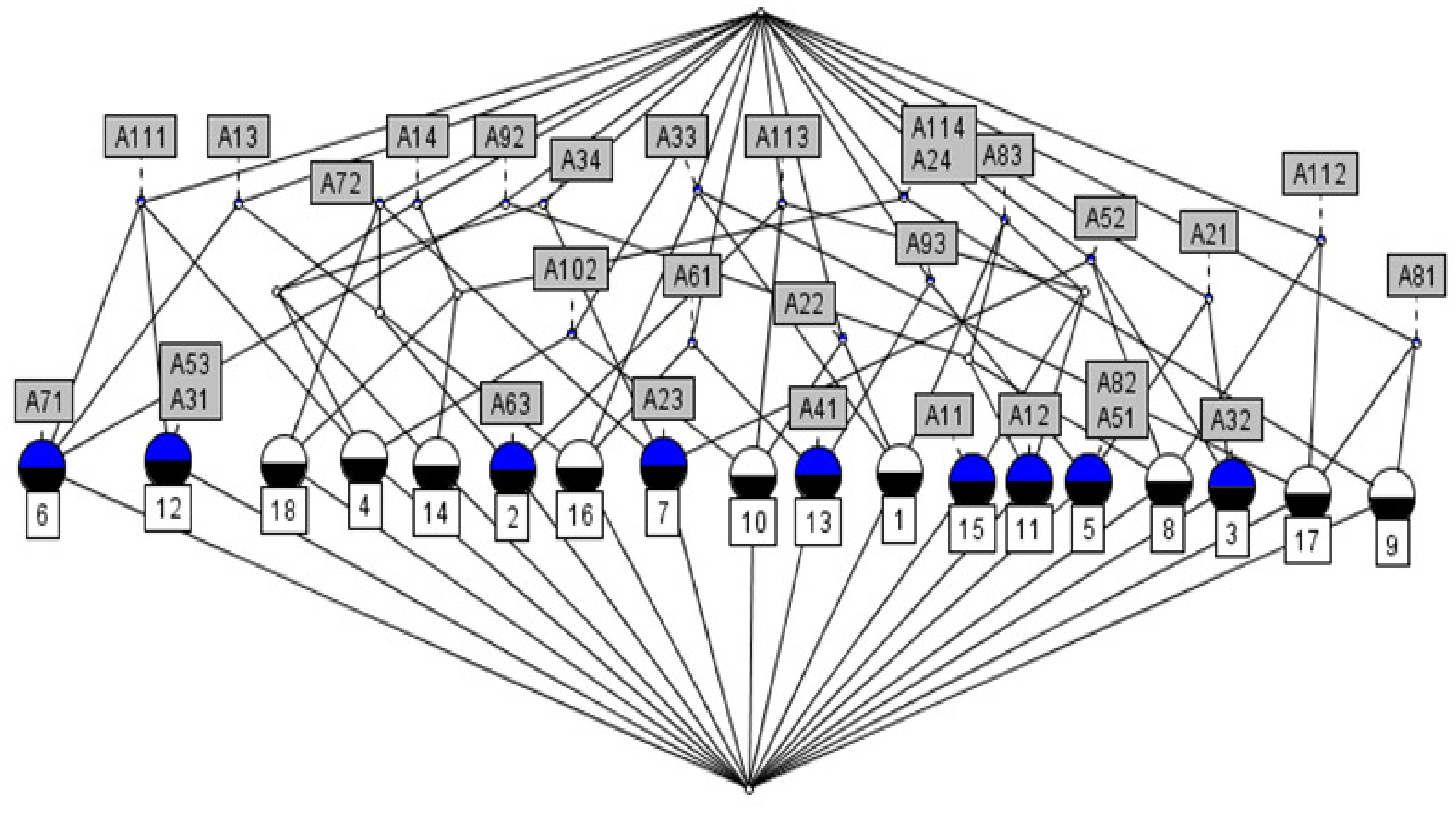

Figure 6 Lattice diagram of the decision class heart failure

From the higher frequency of the implication Table 18, we can find the chief characteristics influencing the decision class heart failure. The most important characteristics due to which heart failure occur are chest pain-asymptomatic and old peak-terrible. The other important characteristics that are equally applicable to this class of heart failure are chest pain-non anginal pain, and thallium scam-fixed defect.

Table 17. Implication set table for decision class heart failure

| **Implication Sets** | | |
|---|---|---|
| 1 < 3 > A114 ==> A24; | 19 < 1 > A33 A72 ==> A13 A61; | 37 < 1 > A34 A111 ==> A14 A102; |
| 2 < 3 > A24 ==> A114; | 20 < 1 > A34 A72 ==> A23 A52; | 38 < 1 > A92 A111 ==> A13 A71; |
| 3 < 1 > A11 ==> A83 A113; | 21 < 1 > A52 A72 ==> A23 A34; | 39 < 1 > A102 A111 ==> A14 A34; |
| 4 < 1 > A12 ==> A83 A92 A113; | 22 < 1 > A61 A72 ==> A13 A33; | 40 < 1 > A33 A112 ==> A81; |
| 5 < 1 > A23 ==> A34 A52 A72; | 23 < 1 > A33 A81 ==> A112; | 41 < 1 > A52 A112 ==> A83 A92; |
| 6 < 1 > A31 ==> A53 A111; | 24 < 1 > A82 ==> A21 A51 A93; | 42 < 1 > A81 A112 ==> A33; |
| 7 < 1 > A32 ==> A21 A52; | 25 < 1 > A22 A83 ==> A33; | 43 < 1 > A83 A112 ==> A52 A92; |
| 8 < 1 > A13 A33 ==> A61 A72; | 26 < 1 > A33 A83 ==> A22; | 44 < 1 > A92 A112 ==> A52 A83; |
| 9 < 1 > A22 A33 ==> A83; | 27 < 1 > A52 A83 ==> A92 A112; | 45 < 1 > A13 A113 ==> A63 A72; |
| 10 < 1 > A41 ==> A61 A93; | 28 < 1 > A13 A92 ==> A71 A111; | 46 < 1 > A22 A113 ==> A102; |
| 11 < 1 > A51 ==> A21 A82 A93; | 29 < 1 > A52 A92 ==> A83 A112; | 47 < 1 > A72 A113 ==> A13 A63; |
| 12 < 1 > A21 A52 ==> A32; | 30 < 1 > A21 A93 ==> A51 A82; | 48 < 1 > A92 A113 ==> A12 A83; |
| 13 < 1 > A34 A52 ==> A23 A72; | 31 < 1 > A61 A93 ==> A41; | 49 < 1 > A102 A113 ==> A22; |
| 14 < 1 > A53 ==> A31 A111; | 32 < 1 > A14 A102 ==> A34 A111; | 50 < 1 > A14 A72 ==> A24 A114; |
| 15 < 1 > A13 A61 ==> A33 A72; | 33 < 1 > A22 A102 ==> A113; | 51 < 1 > A24 A34 A114 ==> A14; |
| 16 < 1 > A33 A61 ==> A13 A72; | 34 < 1 > A34 A102 ==> A14 A111; | 52 < 1 > A24 A72 A114 ==> A14; |
| 17 < 1 > A63 ==> A13 A72 A113; | 35 < 1 > A13 A111 ==> A71 A92; | |
| 18 < 1 > A71 ==> A13 A92 A111; | 36 < 1 > A14 A111 ==> A34 A102; | |

Table 18. Implication relation table for decision class heart failure

| **Superconcept** | $a_{12}$ | $a_{13}$ | $a_{14}$ | $a_{21}$ | $a_{22}$ | $a_{23}$ | $a_{24}$ | $a_{31}$ | $a_{32}$ | $a_{33}$ | $a_{34}$ | $a_{41}$ | $a_{51}$ | $a_{52}$ | $a_{53}$ |
|---|---|---|---|---|---|---|---|---|---|---|---|---|---|---|---|
| **Subconcept** | $a_{92}$,<br>$a_{11,3}$ | $a_{33}$,$a_{61}$<br>$a_{63}$<br>$a_{33}$,$a_{72}$<br>$a_{61}$,$a_{72}$<br>$a_{92}$,$a_{11,1}$<br>$a_{72}$,$a_{11,3}$ | $a_{34}$,$a_{10,2}$<br>$a_{34}$,$a_{11,1}$<br>$a_{10,2}$,$a_{11,1}$<br>$a_{24}$,$a_{34}$,<br>$a_{11,4}$;$a_{24}$<br>$a_{72}$,$a_{11,4}$ | $a_{32}$<br>$a_{51}$<br>$a_{82}$ | $a_{33}$,<br>$a_{83}$;<br>$a_{10,2}$,<br>$a_{11,3}$ | $a_{34}$,<br>$a_{52}$;<br>$a_{34}$,<br>$a_{72}$;<br>$a_{52}$,<br>$a_{72}$ | $a_{11,4}$ *3<br>$a_{14}$,$a_{72}$ | $a_{53}$ | $a_{21}$,<br>$a_{52}$ | $a_{13}$,$a_{61}$<br>$a_{61}$,$a_{72}$<br>$a_{22}$,$a_{83}$<br>$a_{81}$,$a_{11,2}$ | $a_{23}$<br>$a_{14}$,$a_{10,2}$<br>$a_{52}$,$a_{72}$<br>$a_{14}$,$a_{11,1}$<br>$a_{10,2}$,$a_{11,1}$ | $a_{61}$,<br>$a_{93}$ | $a_{21}$,<br>$a_{93}$;<br>$a_{82}$ | $a_{23}$;$a_{32}$<br>$a_{34}$,$a_{72}$<br>$a_{83}$,$a_{11,2}$<br>$a_{92}$,$a_{11,2}$ | $a_{31}$ |
| **Frequency** | 2 | 11 | 12 | 3 | 4 | 6 | 5 | 1 | 2 | 8 | 9 | 2 | 3 | 8 | 1 |
| **Superconcept** | $a_{61}$ | $a_{63}$ | $a_{71}$ | $a_{72}$ | $a_{81}$ | $a_{82}$ | $a_{83}$ | $a_{92}$ | $a_{93}$ | $a_{10,2}$ | $a_{11,1}$ | $a_{11,2}$ | $a_{11,3}$ | $a_{11,4}$ | |
| **Subconcept** | $a_{13}$,<br>$a_{33}$;<br>$a_{41}$<br>$a_{33}$,<br>$a_{72}$ | $a_{72}$,<br>$a_{11,3}$;<br>$a_{13}$,<br>$a_{11,3}$ | $a_{13}$,<br>$a_{11,1}$;<br>$a_{92}$,<br>$a_{11,1}$; | $a_{23}$;$a_{63}$<br>$a_{13}$,$a_{33}$<br>$a_{34}$,$a_{52}$<br>$a_{13}$,$a_{61}$<br>$a_{33}$,$a_{61}$<br>$a_{13}$,$a_{11,3}$ | $a_{33}$,<br>$a_{11,2}$ | $a_{51}$<br>$a_{21}$,<br>$a_{93}$ | $a_{11}$;$a_{12}$<br>$a_{22}$,$a_{33}$<br>$a_{52}$,$a_{92}$<br>$a_{92}$,$a_{11,3}$<br>$a_{92}$,$a_{11,2}$<br>$a_{52}$,$a_{11,2}$ | $a_{12}$;$a_{71}$<br>$a_{52}$,$a_{83}$<br>$a_{13}$,$a_{11,1}$<br>$a_{52}$,$a_{11,2}$<br>$a_{83}$,$a_{11,2}$ | $a_{41}$<br>$a_{51}$<br>$a_{82}$ | $a_{22}$,<br>$a_{11,3}$;<br>$a_{34}$,<br>$a_{11,1}$;<br>$a_{14}$,<br>$a_{11,152}$ | $a_{31}$<br>$a_{53}$<br>$a_{71}$<br>$a_{13}$,$a_{92}$<br>$a_{14}$,$a_{10,2}$<br>$a_{34}$,$a_{10,2}$ | $a_{52}$,$a_{92}$<br>$a_{52}$,$a_{83}$<br>$a_{33}$,$a_{81}$ | $a_{11}$<br>$a_{12}$<br>$a_{63}$<br>$a_{22}$,<br>$a_{10,2}$ | $a_{24}$ *3<br>$a_{14}$,$a_{72}$ | |
| **Frequency** | 5 | 4 | 4 | 12 | 2 | 3 | 12 | 10 | 3 | 6 | 9 | 6 | 5 | 5 | |

## 5.2.4 Decision Class Potential Patient

The number of rules generated using rule generation algorithm due to rough set theory is 17 and is further minimized to 11 by using domain intelligence. This is further validated and minimized to 9 numbers of rules using testing data set. Now we present the context table in Table 19, which converted 9 rules of potential patients representing the attributes obtained from pre process. In Figure 7, we present the lattice diagram of the context Table 19 for the decision class potential patient. The implication set table is presented in Table 20 whereas the implication relation table is given in Table 21.

Table 19. Context table of decision class potential patient

| **Rule** | $a_1$ | | | | $a_2$ | | | | $a_3$ | | | | $a_4$ | | $a_5$ | | | $a_6$ | | | $a_7$ | | $a_8$ | | | $a_9$ | | | $a_{10}$ | | $a_{11}$ | | | |
|---|---|---|---|---|---|---|---|---|---|---|---|---|---|---|---|---|---|---|---|---|---|---|---|---|---|---|---|---|---|---|---|---|---|---|
| | $a_{11}$ | $a_{12}$ | $a_{13}$ | $a_{14}$ | $a_{21}$ | $a_{22}$ | $a_{23}$ | $a_{24}$ | $a_{31}$ | $a_{32}$ | $a_{33}$ | $a_{34}$ | $a_{41}$ | $a_{42}$ | $a_{51}$ | $a_{52}$ | $a_{53}$ | $a_{61}$ | $a_{62}$ | $a_{63}$ | $a_{71}$ | $a_{72}$ | $a_{81}$ | $a_{82}$ | $a_{83}$ | $a_{91}$ | $a_{92}$ | $a_{93}$ | $a_{10,1}$ | $a_{10,2}$ | $a_{11,1}$ | $a_{11,2}$ | $a_{11,3}$ | $a_{11,4}$ |
| [1] | | | × | | | | | | × | | | | | | | | | | | | | | | × | | | | | | × | | | | |
| [2] | | | | × | | | | | | | | | | | | | | | | | | | | × | | | | | | | | × | | |
| [3] | | | | | | | | | | | | | | | | × | | | | | | | × | | | × | | | | | | | | |
| [4] | | | × | | | × | | | | | | | | | | | | | | | | | | | | | | | | | | | × | |
| [5] | | × | | | | | | | | × | | | | | | | | | | | × | | | | | | | | | | × | | | |
| [6] | | | × | | | | | | | | | × | | | | | | | | | | | | | | × | | | | | | | | |
| [7] | | × | | | | | × | | | | | | | | | | | | | | | | × | | | | | | | | | | | |
| [8] | | | | | | | × | | | | | | | | | | | | | | | | | | | × | | | | | | | × | |
| [9] | | | | | | | | × | | | × | | | | | | × | | | | | | | | | | | | | | | | × | |

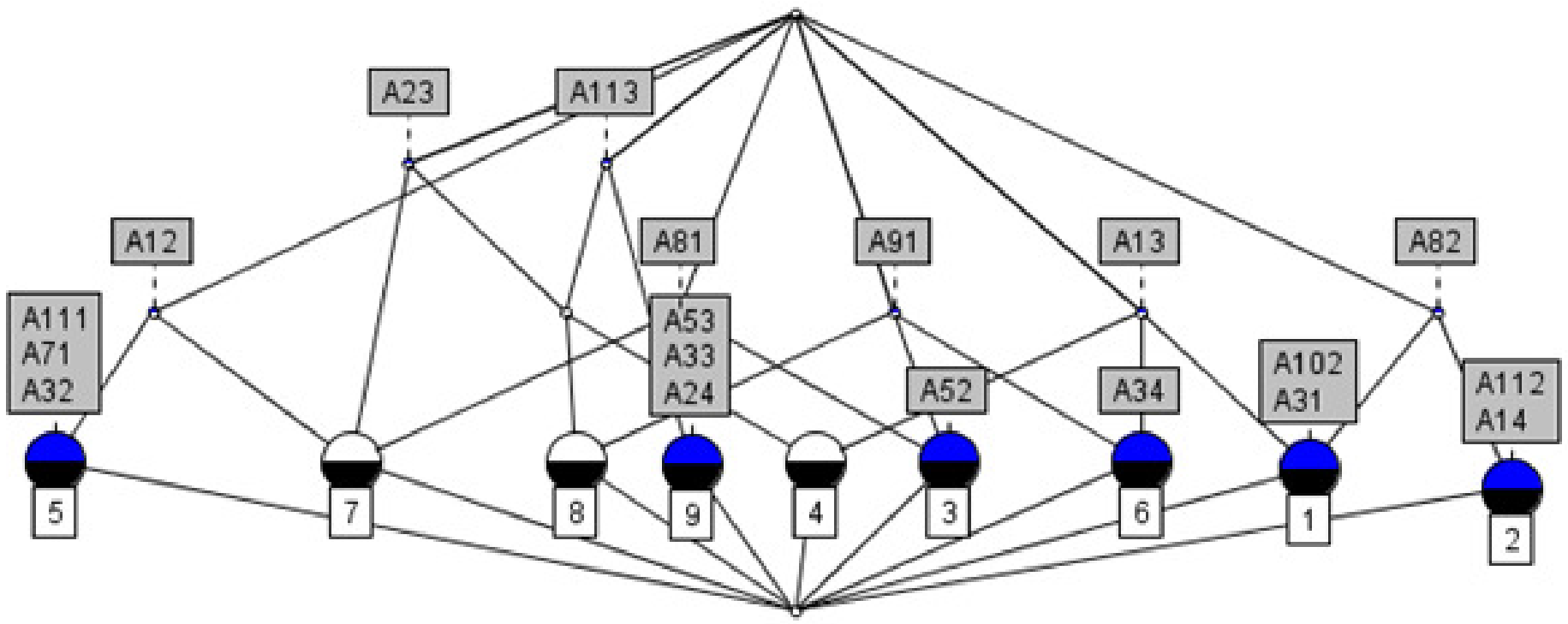

Figure 7 Lattice diagram of the decision class potential patient

From the higher frequency of the implication Table 21 it is observed that the age group old is the potential patients affecting the heart disease. The chief characteristics influencing the decision class is blood pressure-high and the next chief characteristic is chest pain-atypical angina.

Table 20. Implication set table for decision class potential patient

| Implication Sets | | |
|---|---|---|
| 1 < 1 > A14 ==> A82 A112; | 9 < 1 > A52 ==> A81 A91; | 17 < 1 > A81 A91 ==> A52; |
| 2 < 1 > A12 A23 ==> A81; | 10 < 1 > A53 ==> A24 A33 A113; | 18 < 1 > A102 ==> A13 A31 A82; |
| 3 < 1 > A13 A23 ==> A113; | 11 < 1 > A71 ==> A12 A32 A111; | 19 < 1 > A111 ==> A12 A32 A71; |
| 4 < 1 > A24 ==> A33 A53 A113; | 12 < 1 > A12 A81 ==> A23; | 20 < 1 > A112 ==> A14 A82; |
| 5 < 1 > A31 ==> A13 A82 A102; | 13 < 1 > A23 A81 ==> A12; | 21 < 1 > A13 A113 ==> A23; |
| 6 < 1 > A32 ==> A12 A71 A111; | 14 < 1 > A13 A82 ==> A31 A102; | 22 < 1 > A91 A113 ==> A23; |
| 7 < 1 > A33 ==> A24 A53 A113; | 15 < 1 > A13 A91 ==> A34; | |
| 8 < 1 > A34 ==> A13 A91; | 16 < 1 > A23 A91 ==> A113; | |

Table 21. Implication relation table for decision class potential patient

| Superconcept | $a_{12}$ | $a_{13}$ | $a_{14}$ | $a_{23}$ | $a_{24}$ | $a_{31}$ | $a_{32}$ | $a_{33}$ | $a_{34}$ | $a_{52}$ |
|---|---|---|---|---|---|---|---|---|---|---|
| Subconcept | $a_{23}, a_{81}$<br>$a_{11,1}$<br>$a_{71}$<br>$a_{32}$ | $a_{31}$<br>$a_{34}$<br>$a_{10,2}$ | $a_{11,2}$ | $a_{12}, a_{81}$<br>$a_{13}, a_{11,3}$<br>$a_{91}, a_{11,3}$ | $a_{33}$<br>$a_{53}$ | $a_{13}, a_{82}$<br>$a_{10,2}$ | $a_{71}$<br>$a_{11,1}$ | $a_{24}$<br>$a_{53}$ | $a_{13}, a_{91}$ | $a_{81}, a_{91}$ |
| Frequency | 5 | 3 | 1 | 6 | 2 | 3 | 2 | 2 | 2 | 2 |
| Superconcept | $a_{53}$ | $a_{71}$ | $a_{81}$ | $a_{82}$ | $a_{91}$ | $a_{10,2}$ | $a_{11,1}$ | $a_{11,2}$ | $a_{11,3}$ | |
| Subconcept | $a_{24}$<br>$a_{33}$ | $a_{32}$<br>$a_{11,1}$ | $a_{12}, a_{23}$<br>$a_{52}$ | $a_{14}$<br>$a_{31}$<br>$a_{10,2}$<br>$a_{11,2}$ | $a_{34}$<br>$a_{52}$ | $a_{31}$<br>$a_{13}, a_{82}$ | $a_{32}$<br>$a_{71}$ | $a_{14}$ | $a_{24}, a_{33}$<br>$a_{53}$<br>$a_{13}, a_{23}$<br>$a_{23}, a_{91}$ | |
| Frequency | 2 | 2 | 3 | 4 | 2 | 3 | 2 | 1 | 7 | |

### 5.2.5 Decision Class Cardiomyopathy

The number of rules generated using rule generation algorithm due to rough set theory is 17 and is further minimized to 15 by using domain intelligence. This is further validated and minimized to 14 numbers of rules using testing data set. Now we present the context table in Table 22, which converted 14 rules of cardiomyopathy representing the attributes obtained from pre process. In Figure 8, we present the lattice diagram of the context Table 22 for the decision class cardiomyopathy. The implication set table is presented in Table 23 and further the implication relation table is given in Table 24.

Table 22. Context table of decision class cardiomyopathy

| Rule | $a_1$ | | | | $a_2$ | | | | $a_3$ | | | | $a_4$ | | $a_5$ | | | $a_6$ | | | $a_7$ | | $a_8$ | | | $a_9$ | | | $a_{10}$ | | $a_{11}$ | | | |
|---|---|---|---|---|---|---|---|---|---|---|---|---|---|---|---|---|---|---|---|---|---|---|---|---|---|---|---|---|---|---|---|---|---|---|
| | $a_{11}$ | $a_{12}$ | $a_{13}$ | $a_{14}$ | $a_{21}$ | $a_{22}$ | $a_{23}$ | $a_{24}$ | $a_{31}$ | $a_{32}$ | $a_{33}$ | $a_{34}$ | $a_{41}$ | $a_{42}$ | $a_{51}$ | $a_{52}$ | $a_{53}$ | $a_{61}$ | $a_{62}$ | $a_{63}$ | $a_{71}$ | $a_{72}$ | $a_{81}$ | $a_{82}$ | $a_{83}$ | $a_{91}$ | $a_{92}$ | $a_{93}$ | $a_{10,1}$ | $a_{10,2}$ | $a_{11,1}$ | $a_{11,2}$ | $a_{11,3}$ | $a_{11,4}$ |
| [1] | | | | | | × | | | | | | | | | | | × | | | | | | | | | | | | × | | | | | × |
| [2] | | | | | | × | | | | | | | | | | | | | | | | | | × | | | | × | | | × | | | |
| [3] | | | | | | | | | | × | | | | | | | | | | × | | | | | | | | | × | | | | × | |
| [4] | | | | | | | × | | | | | | | | | × | | | | | | | | | | | | | | × | | × | | |
| [5] | | | | | | | | | | | | | × | | | × | | | | × | | | | | | | | × | | | | | | |
| [6] | | × | | | | | | | | | | | | | | | × | | | × | × | | | | | | | | × | | | | | |
| [7] | | | | × | | | | | × | | | | | | | | | | | | | | | | × | | | | | | | | × | |
| [8] | | | | | | | | | | | | | | | | | × | | | | × | | | | | × | | | | | | | | |
| [9] | | | | | | × | | | | × | | | | | | | | | | | | | | | × | | | | | | | | | |
| [10] | | | | × | | × | | | | | | | | | | | | | | | | | | | × | | | | | | | × | | |
| [11] | | | × | | | | | | | | | × | | | | | | | | | | | | | | × | | | | | | | | |
| [12] | | × | | | | | × | | | | | | | | | | | | | | | | × | | | | | | | | | | | |
| [13] | | | | | | | × | | | | | | | | | | | | | | | | | | | × | | | | | | | × | |
| [14] | | | | | | | | × | | | × | | | | | | × | | | | | | | | | | | | | | | | × | |

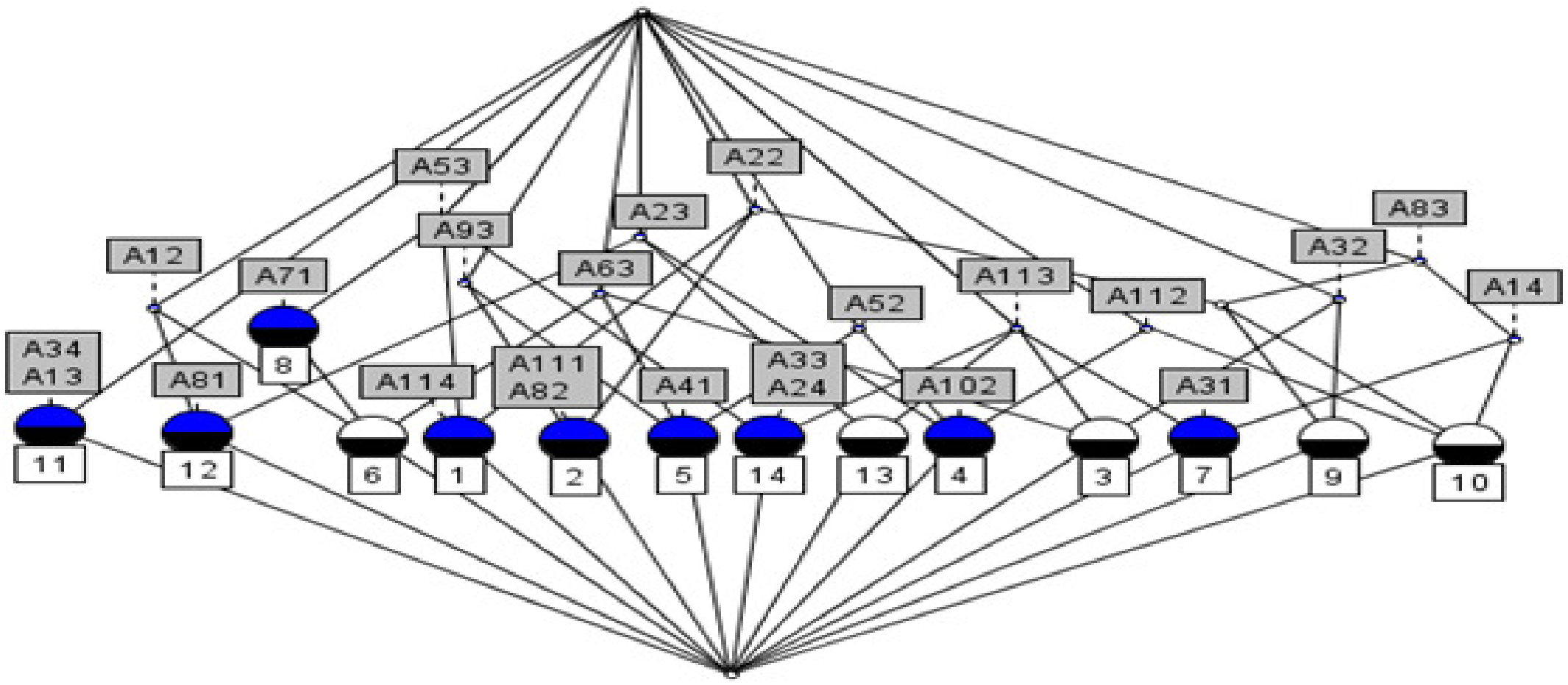

Figure 8 Lattice diagram of the decision class cardiomyopathy

From the higher frequency of the implication Table 24, we obtain that the most important characteristics of cardiomyopathy is maximum heart rate-high and in general it affects to male group. The next important characteristic that leads to this class is ECG-hypertrophy.

Table 23. Implication set table for decision class potential patient

| Implication Sets | | |
|---|---|---|
| 1 < 2 > A14 ==> A83; | 16 < 1 > A53 A63 ==> A12 A71 A101; | 31 < 1 > A22 A112 ==> A14 A83; |
| 2 < 2 > A71 ==> A53; | 17 < 1 > A81 ==> A12 A23; | 32 < 1 > A23 A112 ==> A52 A102; |
| 3 < 1 > A13 ==> A34 A91; | 18 < 1 > A82 ==> A22 A93 A111; | 33 < 1 > A52 A112 ==> A23 A102; |
| 4 < 1 > A12 A23 ==> A81; | 19 < 1 > A32 A83 ==> A22; | 34 < 1 > A83 A112 ==> A14 A22; |
| 5 < 1 > A24 ==> A33 A53 A113; | 20 < 1 > A14 A22 A83 ==> A112; | 35 < 1 > A23 A113 ==> A91; |
| 6 < 1 > A31 ==> A14 A83 A113; | 21 < 1 > A23 A91 ==> A113; | 36 < 1 > A32 A113 ==> A63 A101; |
| 7 < 1 > A22 A32 ==> A83; | 22 < 1 > A53 A91 ==> A71; | 37 < 1 > A53 A113 ==> A24 A33; |
| 8 < 1 > A33 ==> A24 A53 A113; | 23 < 1 > A22 A93 ==> A82 A111; | 38 < 1 > A63 A113 ==> A32 A101; |
| 9 < 1 > A34 ==> A13 A91; | 24 < 1 > A52 A93 ==> A41 A63; | 39 < 1 > A83 A113 ==> A14 A31; |
| 10 < 1 > A41 ==> A52 A63 A93; | 25 < 1 > A63 A93 ==> A41 A52; | 40 < 1 > A91 A113 ==> A23; |
| 11 < 1 > A23 A52 ==> A102 A112; | 26 < 1 > A12 A101 ==> A53 A63 A71; | 41 < 1 > A101 A113 ==> A32 A63; |
| 12 < 1 > A12 A53 ==> A63 A71 A101; | 27 < 1 > A32 A101 ==> A63 A113; | 42 < 1 > A114 ==> A22 A53 A101; |
| 13 < 1 > A12 A63 ==> A53 A71 A101; | 28 < 1 > A53 A71 A101 ==> A12 A63; | 43 < 1 > A22 A53 ==> A101 A114; |
| 14 < 1 > A32 A63 ==> A101 A113; | 29 < 1 > A102 ==> A23 A52 A112; | 44 < 1 > A22 A101 ==> A53 A114; |
| 15 < 1 > A52 A63 ==> A41 A93; | 30 < 1 > A111 ==> A22 A82 A93; | |

Table 24. Implication relation table for decision class cardiomyopathy

| **Superconcept** | $a_{12}$ | $a_{13}$ | $a_{14}$ | $a_{22}$ | $a_{23}$ | $a_{24}$ | $a_{31}$ | $a_{32}$ | $a_{33}$ | $a_{34}$ | $a_{41}$ | $a_{52}$ | $a_{53}$ |
|---|---|---|---|---|---|---|---|---|---|---|---|---|---|
| **Subconcept** | $a_{81}, a_{53}$<br>$a_{63}, a_{53}$<br>$a_{71}, a_{10,1}$ | $a_{34}$ | $a_{31}$<br>$a_{22}, a_{11,2}$<br>$a_{83}, a_{11,2}$<br>$a_{83}, a_{11,3}$ | $a_{38}, a_{32}$<br>$a_{83}, a_{11,1}$<br>$a_{83}, a_{11,2}$<br>$a_{11,4}$ | $a_{91}, a_{11,3}$<br>$a_{52}, a_{11,2}$<br>$a_{10,2}, a_{81}$ | $a_{53}$,<br>$a_{11,3}$,<br>$a_{33}$ | $a_{83}$,<br>$a_{11,3}$ | $a_{63}, a_{11,3}$<br>$a_{10,1}, a_{11,3}$ | $a_{24}$,<br>$a_{53}$,<br>$a_{11,3}$ | $a_{13}$ | $a_{52}, a_{93}$<br>$a_{52}, a_{63}$<br>$a_{63}, a_{93}$ | $a_{10,2}$<br>$a_{23}, a_{11,2}$<br>$a_{63}, a_{93}$ | $a_{71}$ *2<br>$a_{24}, a_{33}$<br>$a_{11,4}$<br>$a_{22}, a_{10,1}$<br>$a_{12}, a_{10,1}$<br>$a_{12}, a_{63}$ |
| **Frequency** | 6 | 1 | 7 | 7 | 6 | 3 | 2 | 4 | 3 | 1 | 6 | 5 | 11 |
| **Superconcept** | $a_{63}$ | $a_{71}$ | $a_{82}$ | $a_{83}$ | $a_{91}$ | $a_{93}$ | $a_{10,1}$ | $a_{10,2}$ | $a_{11,1}$ | $a_{11,2}$ | $a_{11,3}$ | $a_{11,4}$ | |
| **Subconcept** | $a_{41}, a_{12}$,<br>$a_{53}, a_{52}$,<br>$a_{93}, a_{12}$,<br>$a_{10,1}, a_{32}$,<br>$a_{10,1}, a_{32}$,<br>$a_{11,3}, a_{10,1}$,<br>$a_{11,3}, a_{53}$,<br>$a_{71}, a_{10,1}$ | $a_{12}, a_{53}$<br>$a_{12}, a_{63}$<br>$a_{12}, a_{10,1}$<br>$a_{53}, a_{63}$<br>$a_{53}, a_{91}$ | $a_{11,1}$<br>$a_{22}, a_{93}$ | $a_{14}$ *2<br>$a_{31}$<br>$a_{22}, a_{32}$ | $a_{23}, a_{11,3}$<br>$a_{34}$ | $a_{52}, a_{63}$<br>$a_{82}, a_{11,1}$ | $a_{12}, a_{53}$<br>$a_{12}, a_{63}$<br>$a_{32}, a_{63}$<br>$a_{53}, a_{63}$<br>$a_{32}, a_{11,3}$<br>$a_{63}, a_{11,3}$<br>$a_{11,4}$<br>$a_{22}, a_{53}$ | $a_{23}, a_{11,2}$<br>$a_{52}, a_{11,2}$ | $a_{82}$<br>$a_{22}, a_{93}$ | $a_{10,2}, a_{23}$<br>$a_{52}, a_{14}$<br>$a_{22}, a_{83}$ | $a_{23}$,<br>$a_{91}$ | $a_{22}$,<br>$a_{53}$,<br>$a_{22}$,<br>$a_{10,1}$ | |
| **Frequency** | 16 | 10 | 3 | 5 | 3 | 4 | 15 | 4 | 3 | 6 | 2 | 4 | |

## 6. Conclusion

In this study, the rule generation algorithm of rough set theory generates 91 rules. This is further minimized to 72 candidate rules with the help of domain intelligence and is further minimized to 65 rules by validation process and threshold value. Further these suitable rules are explored to identify the chief characteristics affecting the relationship between heart disease and its attributes by using formal concept analysis. This helps the decision maker a priori detection of the heart disease. The chief characteristics of hypertensive heart disease are old peak ($a_{83}$), exercise ($a_{71}$); the other characteristics include chest pain ($a_{12}$), blood pressure ($a_{23}$), and maximum heart rate ($a_{63}$).The chief characteristics of coronary heart disease are thallium scan ($a_{93}$), exercise ($a_{71}$); the other characteristics include chest pain ($a_{12}$), and blood pressure ($a_{24}$). The chief characteristics of heart failure are chest pain ($a_{14}$), exercise ($a_{72}$), and old peak ($a_{83}$); the other characteristics include chest pain ($a_{13}$), and thallium scan ($a_{92}$).The chief characteristics of patient that may lead to heart disease are blood pressure ($a_{23}$), and chest pain ($a_{12}$), where the targeted age group is old. Finally, the chief characteristics of cardiomyopathy are maximum heart rate ($a_{63}$), and ECG ($a_{53}$), where the targeted group in general is male. We believe that, formal concept analysis can be used to find furthermore information regardless of the type of rule based soft computing. We also believe that the proposed model is a useful method for decision makers.

**Authors**

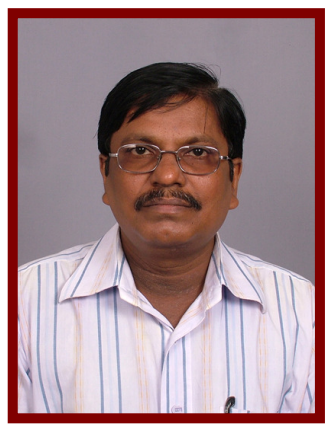

**B K Tripathy**, a senior professor in the school of computing sciences and engineering, VIT University, at Vellore, India. He has been awarded with Gold Medals both at graduate and post graduate levels of Berhampur University, India. Also, he has been awarded with the best post graduate of the Berhampur University. He has received national scholarship, UGC fellowship, SERC visiting fellowship and DOE (Govt. of India) scholarship at various levels of his career. He has published more than hundred technical papers in various international journals, conferences, and Springer book chapters. He has produced ten Ph. Ds under his supervision. He is associated with many professional bodies like IEEE, IRSS, WSEAS, AISTC, ISTP, CSI, AMS, and IMS. His name also appeared in the editorial board of several international journals like CTA, ITTA, AMMS, IJCTE, AISS, AIT, and IJPS. Also, he is a reviewer of international journals like Mathematical Reviews, Information Sciences, Analysis of Neural Networks, Journal of Knowledge Engineering, Mathematical Communications, IJET, IJPR and Journal of Analysis. His name has come out in Marquis who's who and in International Biographical Centre, London. His research interest includes fuzzy sets and systems, rough sets and knowledge engineering, data clustering, social network analysis, soft computing and granular computing.

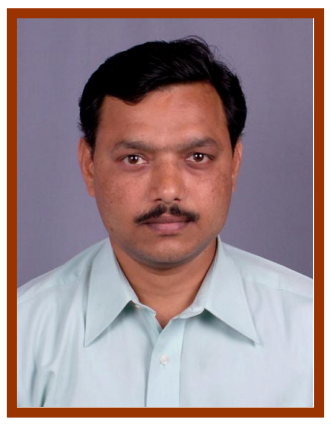

**D P Acharjya** received the M. Tech. degree in computer science from Utkal University, India in 2002; M. Phil. from Berhampur University, India; and M. Sc. from NIT, Rourkela, India. He has been awarded with Gold Medal in M. Sc. He is a research student of Berhampur University, India. Currently, he is an Associate Professor in the school of computing sciences and engineering, VIT University, Vellore, India. His 15 years of experience in higher education include as a faculty member, reviewer, academic councillor, academic guide and member of board of studies. He has authored many national and international journal papers and three books entitled Fundamental Approach to Discrete Mathematics, 2nd Ed, New-Age International Pvt. Ltd, New Delhi, India (2009); Computer Based on Mathematics, Laxmi Publications, New Delhi, India (2007); and Theory of Computation, MJP Publishers, Chennai, India (2010) to his credit. He is associated with many professional bodies CSI, ISTE, IMS, AMTI, ISIAM, OITS, IACSIT, CSTA, IEEE and IAENG. He was founder secretary of OITS Rourkela chapter. His current research interests include rough sets, formal concept analysis, knowledge representation, granular computing, data mining and business intelligence.

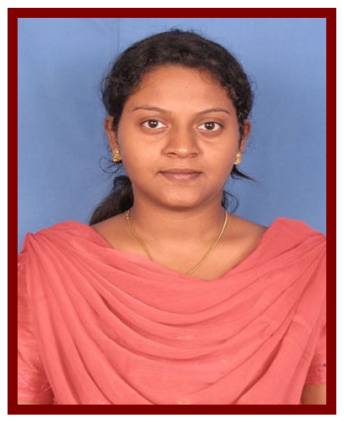

**V. Cynthya**, received the B. E. in Computer Science from Karunya University, Coimbatore, India in 2009. She is a M. Tech. (CSE) final year student of VIT University, Vellore, India. She has keen interest in teaching and applied research. Her research interests include rough sets, granular computing, formal concepts and knowledge mining.